\documentclass[journal]{IEEEtran}
\usepackage{cite}
\usepackage{hyperref}
\usepackage{comment}

\ifCLASSINFOpdf
   \usepackage[pdftex]{graphicx}
\else
   \usepackage[dvips]{graphicx}
\fi
\usepackage{amsmath,amssymb}
\DeclareMathOperator{\Tr}{Tr}
\usepackage{makecell}
\usepackage{multirow}
\usepackage{multicol}
\usepackage{xspace}
\usepackage{booktabs}
\usepackage{caption}
\usepackage[dvipsnames]{xcolor}
\usepackage{cuted}

\usepackage{amsthm}

\newtheorem{proposition}{Proposition}
\usepackage{dsfont}
\usepackage{pifont}
\newcommand{\newcheckmark}{\textrm{\ding{51}}}%
\newcommand{\newcrossmark}{\textrm{\ding{55}}}%
\usepackage{wrapfig}

\makeatletter
\DeclareRobustCommand\onedot{\futurelet\@let@token\@onedot}
\def\@onedot{\ifx\@let@token.\else.\null\fi\xspace}

\def\eg{\emph{e.g}\onedot} 
\def\ie{\emph{i.e}\onedot}

\makeatother

\usepackage{url}

\begin{document}
%
\title{Highly Efficient 3D Human Pose Tracking from Events with Spiking Spatiotemporal Transformer}
%
%
%

\author{Shihao~Zou,
        Yuxuan~Mu,
        Wei~Ji,
        Zi-An~Wang,
        Xinxin~Zuo,
        Sen~Wang,
        Weixin~Si,
        Li~Cheng
\thanks{S. Zou and Z. Wang are with Shenzhen Institutes of Advanced Technology, Chinese Academy of Sciences, Shenzhen, Guangdong, China. (E-mail: sh.zou@siat.ac.cn, za.wang@siat.ac.cn).}
\thanks{W. Si is with Faculty of Computer Science and Control Engineering, Shenzhen University of Advanced Technology,  Shenzhen, Guangdong, China. (E-mail: wx.si@siat.ac.cn).}
\thanks{Y. Mu, W. Ji, S. Wang and L. Cheng are with the Department
of Electrical and Computer Engineering, University of Alberta, AB, Canada, T6G 1H9. (E-mail: ymu3@ualberta.ca, wji3@ualberta.ca, sen9@ualberta.ca, lcheng5@ualberta.ca).}
\thanks{X. Zuo was with the Department of Electrical and Computer Engineering, University of Alberta, and now with the Department of Electrical and Computer Engineering, Concordia University. (E-mail: xinxin.zuo@concordia.ca)}
\thanks{Weixin Si and Li Cheng are the corresponding authors for this paper.}
}

%
%

\markboth{Journal of \LaTeX\ Class Files,~Vol.~14, No.~8, August~2015}%
{Shell \MakeLowercase{\textit{et al.}}: Bare Demo of IEEEtran.cls for IEEE Journals}
%



\maketitle

\begin{abstract}
Event camera, as an asynchronous vision sensor capturing scene dynamics, presents new opportunities for highly efficient 3D human pose tracking. 
Existing approaches typically adopt modern-day Artificial Neural Networks (ANNs), such as CNNs or Transformer, where sparse events are converted into dense images or paired with additional gray-scale images as input. 
Such practices, however, ignore the inherent sparsity of events, resulting in redundant computations, increased energy consumption, and potentially degraded performance. 
Motivated by these observations, we introduce the first sparse Spiking Neural Networks (SNNs) framework for 3D human pose tracking based solely on events. Our approach eliminates the need to convert sparse data to dense formats or incorporate additional images, thereby fully exploiting the innate sparsity of input events. 
Central to our framework is a novel Spiking Spatiotemporal Transformer, which enables bi-directional spatiotemporal fusion of spike pose features and provides a guaranteed similarity measurement between binary spike features in spiking attention.
Moreover, we have constructed a large-scale synthetic dataset, SynEventHPD, that features a broad and diverse set of 3D human motions, as well as much longer hours of event streams. 
Empirical experiments demonstrate the superiority of our approach over existing state-of-the-art (SOTA) ANN-based methods, requiring only 19.1\% FLOPs and 3.6\% energy cost. Furthermore, our approach outperforms existing SNN-based benchmarks in this task, highlighting the effectiveness of our proposed SNN framework. The dataset will be released upon acceptance, and code can be found at \href{https://github.com/JimmyZou/HumanPoseTracking\_SNN}{https://github.com/JimmyZou/HumanPoseTracking\_SNN}.
\end{abstract}


%
\IEEEpeerreviewmaketitle

\section{Introduction}\label{sec:introduction}

Human pose tracking in 3D~\cite{gai2023spatiotemporal,zhou2024dual,tang2024ftcm,shi2024identify,liu2024event,humanMotionKanazawa19,kocabas2019vibe} has attracted increasing research attentions in recent years within video technology due to its broad applications in areas such as human-computer interaction, video-based analysis, and virtual/augmented reality. While most current research efforts have been focused on the frame-based RGB cameras, event cameras~\cite{gallego2019event}, as an emerging vision sensor, present new opportunities for developing more efficient systems in this task. As a biologically-inspired vision system, event cameras are considerably different from the standard frame-based cameras. Leveraging their distinct sparse, asynchronous, and independent address-event representation, event cameras capture object motions as a sparse set of pixel-wise events at remarkably low energy costs. This innovative imaging paradigm has sparked a multitude of research efforts in event-based vision tasks, such as semantic segmentation~\cite{zhou2023spikingformer}, object tracking~\cite{zhang2021object,zhang2022spiking}, recognition~\cite{kim2022ev,fang2021deep}, 3D reconstruction~\cite{rebecq2018emvs,zhang2022discrete}, robotics and autonomous driving~\cite{gallego2019event}.

Recent data-driven approaches have shown their potential in 3D human pose estimation from event cameras~\cite{calabrese2019dhp19,xu2020eventcap,zou2021eventhpe,rudnev2021eventhands,scarpellini2021lifting}. Early efforts~\cite{calabrese2019dhp19,scarpellini2021lifting} convert events to dense images and feed into ANN models to estimate 2D human poses, which are then lifted to 3D. There is also effort~\cite{rudnev2021eventhands} in applying ANN models for static hand pose estimation from events. In contrast, EventCap~\cite{xu2020eventcap} adopts a different strategy, taking a stream of 2D events and its corresponding gray-scale video from the same viewpoint as input to track 3D poses. EventHPE~\cite{zou2021eventhpe} further reduces the dependence on gray-scale video. It uses only the initial gray-scale frame to extract the starting pose, and then relies solely on events for subsequent pose tracking.
Unfortunately, existing methods either adhere to conventional process of converting events into dense images for ANN input~\cite{calabrese2019dhp19,rudnev2021eventhands,scarpellini2021lifting}, or require additional gray-scale images for initial poses extraction using pre-trained ANN models~\cite{xu2020eventcap,zou2021eventhpe}. These approaches usually overlook the inherent sparsity of event signals, as reflected in the FLOPs and energy consumption reported in Tab.~\ref{tab:pose-estimation}, thereby compromising energy efficiency in event-driven applications. Consequently, the full potential of 3D human pose tracking based solely on events remains largely unexploited.

Meanwhile, ANNs, such as ResNet~\cite{he2016deep} and Transformer~\cite{vaswani2017attention}, have demonstrated their competence in various event-based vision tasks~\cite{rudnev2021eventhands,wang2020eventsr,fang2021deep,fang2021incorporating,kim2022ev,zhang2021object,sun2022ess,gehrig2019end,gehrig2020video}. However, compared with dense RGB or gray-scale images, event streams are spatiotemporally much sparser, resulting in a growing interest in the development of SNNs to efficiently process event signals. Unlike ANNs, SNNs employ spiking neurons to replicate the event generation process, and thus can effectively bypass the unnecessary computations of non-spiking neurons, where no events are generated. Recent efforts have demonstrated the effectiveness of SNNs, either by converting ANNs to SNNs~\cite{rueckauer2017conversion,deng2021optimal}, or training SNNs from scratch~\cite{fang2021deep,fang2021incorporating,li2021differentiable,yao2022glif,yao2023attention,zhou2022spikformer,zhou2023spikingformer}. 
But these models are primarily validated on high-level vision classification tasks, which are fundamentally different from the task of 3D human pose tracking. 
Instead of assigning a single categorical label, our task is designed to regress a sequence of fine-grained pose values over time. Therefore, the effectiveness of an SNN approach in the task of human pose tracking has not yet been fully investigated.

Motivated by these observations, this paper focuses on a relatively new task: 3D human pose tracking using only events, approached solely through SNNs to fully exploit the innate sparsity of events data. 
This is a challenging task as spiking trains in existing SNN models~\cite{rueckauer2017conversion,fang2021deep} are typically unfolded over time, preserving only one-directional temporal dependency. This may lead to insufficient pose information during the early time intervals. For instance, when the person is stationary at the beginning, few events can be captured for accurate early-time pose tracking.
Although recently proposed Spiking Transformers~\cite{zhou2022spikformer,zhou2023spikingformer} offer a potential solution of spiking attention for global feature fusion, their reliance on dot-product for computing attention scores between binary spike features can lead to an ill-defined similarity measurement and the risk of feature vanishing, as illustrated in Fig.~\ref{fig:spiking-transformer}~(c). The other Spiking Transformers introduced in~\cite{zhang2022spiking,zhang2022spikedepth} are actually not SNN models, but either dense ANNs or a mixed framework of ANNs and SNNs, which lack the energy efficiency advantage typically associated with SNNs.

Therefore, we introduce a novel Spiking Spatiotemporal Transformer, an SNN approach that facilitates bi-directional spatiotemporal fusion of spike pose features, enabling the propagation of human pose information, especially to early time intervals. Furthermore, we propose using normalized Hamming similarity as a guaranteed similarity measurement between binary spike features in spiking attention. As demonstrated in Proposition~\ref{proposition:hamming-similarity}, this method closely approximates cosine similarity between real-valued features, providing a feasible mapping of the architecture from standard Transformer~\cite{vaswani2017attention} to Spiking Transformer. 

To summarize, there are three main contributions in our work. (i) We introduce the first SNN approach for the task of 3D human pose tracking solely from events. Our approach fully exploits the innate sparsity in events for minimal computational and energy demands in this task. At its core, we propose a novel Spiking Spatiotemporal Transformer, which addresses the one-directional temporal dependency problem in SNNs and also provides a guaranteed similarity measurement between binary spike features in spiking attention. (ii) We provide a large-scale dataset, SynEventHPD, to facilitate the research in event-based 3D human pose tracking, with a total size of 45.72 hours event streams -- more than 10 times larger than MMHPSD~\cite{zou2021eventhpe}, the existing largest event-based human pose dataset. (iii) Extensive empirical experiments show that our proposed approach outperforms SOTA ANN-based methods~\cite{zou2021eventhpe,carion2020end} while using less than 19.1\% FLOPs and 3.6\% energy cost. It also surpasses existing SNN models~\cite{fang2021deep,yao2023attention,zhou2022spikformer,zhou2023spikingformer} in 3D human pose tracking, showcasing the effectiveness of our proposed SNN framework.

Our preliminary work was published in~\cite{zou2021eventhpe}. This paper extends our preceding effort in a number of aspects: 
\begin{itemize}
    \item Compared with our preceding work that requires additional dense images, this research seeks to develop a highly efficient 3D human pose tracking system that relies exclusively on sparse event data. Event cameras capture only the sparse events that emphasize human motion, while traditional full-frame images often contain redundant information, such as backgrounds and other static objects, as illustrated in Fig.~\ref{fig:vis-pose}. This inherent sparsity of events offers the potential to develop a highly efficient human pose tracking system.
    
    \item Departing from the dense ANNs used in our previous work, we introduce the first SNN approach for this task. SNNs closely mimic the event generation process, motivating our investigation into their effectiveness for human pose tracking and enabling us to fully exploit the sparsity of event signals. To enhance our approach's performance, we propose a Spiking Spatiotemporal Transformer, which addresses the issues of one-directional temporal dependency in SNNs and the ambiguities in similarity measurement within spiking attention.

    \item Expanding upon our previous MMHPSD dataset, we introduce SynEventHPD, a large-scale dataset synthesizing events from four popular motion capture datasets. This dataset significantly increases the variety of motions compared to the earlier MMHPSD dataset~\cite{zou2021eventhpe}, as illustrated in Fig.~\ref{fig:poses-tsne}. With a total duration of 45.72 hours of event streams (over ten times larger than MMHPSD).
    
    \item Our proposed SNN approach outperforms previous event-based ANN-based methods~\cite{zou2021eventhpe,carion2020end}, utilizing only 19.1\% FLOPs and 3.6\% energy cost, underscoring the efficiency of SNNs in event-based human pose tracking.
\end{itemize}

\section{Related Work}
\label{sec:related-work}
\textbf{Human pose estimation} from RGB or depth images has been a popular topic in computer vision in recent years. Prior to the deep learning era, research efforts are mainly based on random forest or dictionary learning~\cite{zhou2016sparseness}. Considering the significant performance boosts brought by deep learning, recent efforts in human pose estimation either directly regress 3D pose from images or lift 2D pose estimation to 3D~\cite{wang20193d}. This trend is further fueled by the development of SMPL~\cite{loper2015smpl}, a parametric human shape model of low-dimensional statistical representation. HMR~\cite{kanazawa2018end} is the first such effort in applying convolutional neural networks for human shape recovery from single images, that produces excellent results. There are also a number of recent efforts to exploit temporal information in inferring human poses and shapes from videos, including e.g. temporal constraints~\cite{kocabas2019vibe}, dynamic cameras or event signals~\cite{xu2020eventcap,zou2021eventhpe}. Recent research~\cite{gai2023spatiotemporal} presents a spatiotemporal learning transformer for RGB video-based human pose estimation, whereas prior work~\cite{zhou2024dual} introduces a dual-path transformer network that captures multiple human contexts and motion details for spatial-temporal modeling. Furthermore, a frequency-temporal collaborative module~\cite{tang2024ftcm} has been designed to encode cross-pose correlations in both frequency and temporal domains, aiming to effectively model global and local dependencies with a more lightweight architecture. In contrast, our method centers on event-based human pose tracking, leveraging the sparsity of event signals and constructing a highly efficient SNN framework.

\textbf{Event cameras}~\cite{gallego2019event}, as a novel brain-inspired technology resembling silicon retinas, differ significantly from conventional frame-based imaging sensors like RGB or Time-of-Flight cameras, particularly in their asynchronous and independent \textit{address-event} representation. The output of an event camera consists of a sequence of ``events'' or ``spikes''. Each readout event is represented as a tuple $(\mathbf{x}, t, p)$. An \textit{event} occurs when the brightness change at pixel position $\mathbf{x}$ (\textit{address}) exceeds a preset threshold at time $t$. The binary polarity status $p$ indicates whether brightness has increased or decreased. Unlike frame-based cameras that densely capture pixel values at a fixed frame rate, event cameras record intensity changes asynchronously and independently, allowing for higher temporal resolution during motion. Additionally, since the output consists of a much sparser stream of motion events, event cameras typically consume considerably less energy during operation.

\textbf{Event-based vision} applications have witnessed a substantial increase in recent years, including camera pose estimation~\cite{gallego2017event}, feature tracking~\cite{gehrig2018asynchronous}, optical flow~\cite{hagenaars2021self}, multi-view stereo~\cite{zhang2022discrete}, and pose estimation~\cite{rudnev2021eventhands}, motion deblurring~\cite{sun2022event}, image restoration and super-resolution~\cite{wang2020eventsr}, image classification~\cite{fang2021deep}, object recognition~\cite{kim2022ev} and tracking~\cite{zhang2021object,zhang2022spiking}, semantic segmentation~\cite{sun2022ess}, events from and to video~\cite{gehrig2019end,gehrig2020video}, depth estimation~\cite{zhang2022spikedepth}, among others. Specifically, recent research~\cite{shi2024identify} investigates the optical characteristics and event-triggering mechanisms of various light interference patterns and introduces an event-based removal method to eliminate interference signals in event streams across both static and dynamic scenes. Another work~\cite{liu2024event} presents an event-driven monocular depth estimator with recurrent transformers, integrating a recursive mechanism to exploit the rich temporal information embedded in event data. Unlike these ANN-based approaches, we propose an efficient SNN-based transformer that fully capitalizes on the inherent sparsity of the input events.

As for human pose estimation, DHP19~\cite{calabrese2019dhp19} is perhaps the first effort in engaging CNNs for event camera based human pose estimation. 
EventCap~\cite{xu2020eventcap} aims to capture 3D motions from both events and gray-scale images provided by an event camera. This work starts with a pre-trained CNN-based 3D pose estimation module that takes a sequence of low-frequency gray-scale images as input; the estimated poses are then used as the initial state to infill the intermediate poses for high-frequency motion capture with the constraint of detected event trajectories by~\cite{gehrig2018asynchronous} and the silhouette information gathered from the events. These methods, however, require full access to the corresponding gray-scale images as co-input. EventHPE~\cite{zou2021eventhpe} reduces this demand by the milder need of only a single gray-scale image of the starting pose. To do this, a dedicated CNN module is trained to infer optical flow by self-supervised learning, which is used alongside with an event stream as input to track 3D poses. Compared with these existing efforts~\cite{xu2020eventcap,zou2021eventhpe}, our approach requires only the events as input, where a novel SNN-based framework is introduced for developing a highly efficient system.

\textbf{Event-based datasets} are crucial for data-driven approaches to attain their satisfactory performance. Unfortunately, existing benchmark datasets are mostly based on conventional RGB or depth cameras -- thus are infeasible to be directly used in event-based tasks, given the fundamental differences between event and conventional cameras. This have motivated a variety of event-based datasets released in recent years, including CIFAR10-DVS~\cite{li2017cifar10} and ES-ImageNet~\cite{lin2021imagenet} for object classification, DSEC-Semantic~\cite{sun2022ess} for semantic segmentation, FE108~\cite{zhang2021object} for object tracking. As for event-based human pose estimation, DHP19 dataset~\cite{calabrese2019dhp19} is the earliest one but has limited amount of events data and lacks pose variety. Our preceding dataset is MMPHSPD~\cite{zou2021eventhpe} with event camera and 3 other imaging modalities. Although the dataset provides more than 4.5 hours event stream and 21 different types of action, it still suffers from pose variety because of in-house constraint environment. Our work further augments MMPHSPD dataset by synthesizing events data from several human motion capture datasets, and finally provides a large-scale dataset with a rich variety of poses for event-based human pose tracking.

\textbf{Spiking neural networks} have been an emerging learning framework. Spiking neuron, the basic element in SNNs, works by imitating the transmitting mechanism in mammalian’s visual cortex~\cite{gallego2019event}. A spiking neuron maintains a membrane potential, which could be changed only when spikes (\textit{events}) are received from its connected preceding neurons. A spike is produced when the neuronal potential exceeds a preset threshold. Different from the neuron in traditional ANNs, no output would be produced by spiking neurons as long as their potentials are below  threshold, thus no computation is taken place -- the root cause of the remarkable efficiency and sparsity of SNNs when comparing to the dense and computational-heavy ANNs. 

\textbf{Training large-scale SNNs} from scratch presents a significant challenge. To address the non-differentiable issue of neuronal spiking function, one branch of works focus on converting trained ANNs to SNNs~\cite{rueckauer2017conversion,deng2021optimal}. It is worth noting that only in the realm of classification tasks, excellent results have been demonstrated by the SNNs methods; the performance is still lacking in fine-grained regression tasks and specifically in human pose estimation. Meanwhile, another branch of works focus on training SNNs from scratch, often by following the back-propagation through time (BPTT) framework and applying surrogate derivatives~\cite{li2021differentiable} to approximate the gradient of neuronal spiking function. This line of works have delivered impressive performance in classification tasks~\cite{fang2021deep,li2021differentiable,yao2023attention,zhou2022spikformer} as well as regression tasks~\cite{hagenaars2021self}. There have also been efforts~\cite{yao2021temporal,zhang2022spiking} in proposing mixed frameworks blending SNNs and ANNs, in order to maintain a good balance in efficiency and performance for event-based tasks. Our proposed work can be regarded as an attempt along this direction, with its focus on the fine-grained regression task of full-body pose tracking. 

\begin{figure*}[t]
    \centering
    \includegraphics[width=\textwidth]{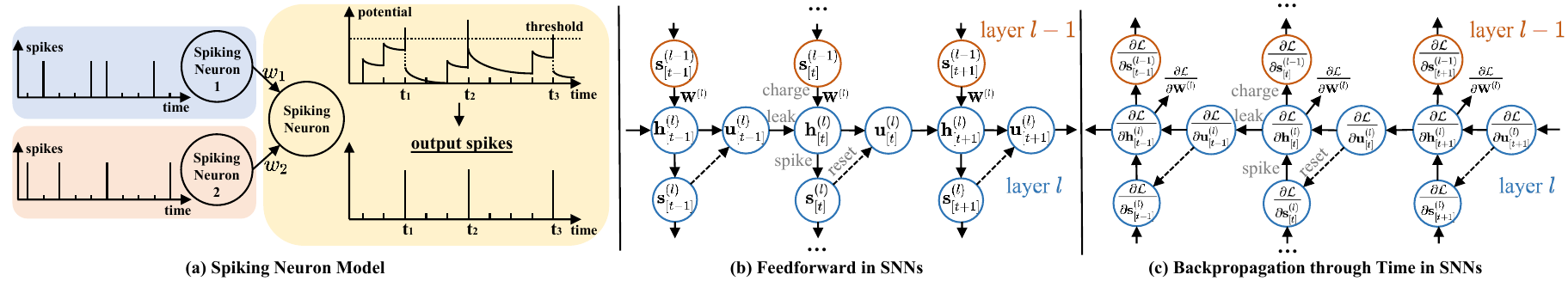}
    \caption{\textbf{(a) Spiking neuron model.} A LIF spiking neuron maintains a membrane potential and modifies it when receiving spiking trains from its connected neurons. The neuron will generate output spikes when its potential exceeds a threshold and then reset the potential. \textbf{(b) Feedforward in SNNs.} This process includes potential leaking and charging, neuron spiking and potential resetting. Feed-forwarding typically rolls over time and propagates from layer $l-1$ to layer $l$. \textbf{(c) Backpropagation Through Time in SNNs.} The gradients are normally computed through time and then back-propagated from layer $l$ to layer $l-1$.}
    \label{fig:neuron_model}
\end{figure*}

\textbf{Spiking Transformer} has emerged very recently as a new SNNs architecture. To avoid confusion, it is important to clarify that the spiking transformers presented in~\cite{zhang2022spiking,zhang2022spikedepth} are not SNN-based transformers, but ANN-based or mixed models. The two recent works~\cite{yao2023attention,zhou2022spikformer} are most related to our proposed spiking spatiotemporal transformer. In MA-SNN~\cite{yao2023attention}, multi-dimensional attention are proposed in an SNNs framework, yet this attention is instead based on real values of membrane potentials, thus not exploring effective attention design for binary spike tensors. In contrast, the spatiotemporal attention mechanism in our model is based on binary spike tensors, which is realised by the proposed Hamming similarity score. It is also very different from the approach taken by Spikformer~\cite{zhou2022spikformer} and Spikingformer~\cite{zhou2023spikingformer}, where the ill-defined dot-product is directly adopted to compute the similarity score of binary spike vectors.

\textbf{Spatiotemporal learning.} Spatial-temporal feature fusion is essential for video-based vision tasks. Earlier work, DBLRNet~\cite{zhang2018adversarial}, applies modified 3D convolutions to both spatial and temporal domains for video deblurring, while ESTINet~\cite{zhang2022enhanced} combines deep residual networks with convolutional LSTMs to model spatial and temporal correlations for video deraining. A self-supervised approach~\cite{liang2022self} simultaneously tackles three continuity-related pretext tasks, enhancing motion (spatial) and context (temporal) representation learning for downstream applications such as action recognition, localization, and video retrieval. Recent studies~\cite{gu2024context, zhou2024dual} employ two-branch transformer designs for decomposed spatial-temporal feature learning in context-guided video grounding, while others~\cite{gai2023spatiotemporal} apply global attention for more precise human pose estimation. Unlike these ANN-based methods, our approach leverages SNNs, offering a solution for sparse spatiotemporal learning.

\section{Preliminary}
\label{sec:preliminary} 
\textbf{Spiking neuron model.} The commonly-used leaky integrate and fire (LIF) model is a fundamental unit in SNNs. As illustrated in Fig.~\ref{fig:neuron_model}~(a), a LIF neuron maintains a membrane potential $u_{[t]}$ with a leaky constant $\tau$, which may be modified only when new spiking trains $X_{[t]}$ are received from its connected neurons in $T$ time steps. When the potential exceeds a predetermined threshold, $V_{\text{th}}$, the neuron outputs a spike $s_{[t]}$ and undergoes a soft reset~\cite{zhou2022spikformer}, decreasing its potential by $V_{\text{th}} - u_{\text{rest}}$. The model is formulated as follows:
\begin{gather}
    h_{[t]} = u_{[t-1]} - \frac{1}{\tau}(u_{[t-1]} - u_{\text{rest}}) + X_{[t]}, \nonumber\\
    s_{[t]} = \Theta (h_{[t]} - V_{\text{th}}),\quad u_{[t]} = h_{[t]} - (V_{\text{th}} - u_{\text{rest}})s_{[t]}, \nonumber
\end{gather}
where $\Theta$ is the Heaviside step function:
\begin{equation}
    \Theta(h_{[t]} - V_{\text{th}}) =
    \begin{cases}
    1, & \text{if } h_{[t]} - V_{\text{th}} \geq 0\\
    0, & \text{otherwise}
    \end{cases} \nonumber
\end{equation}

\textbf{Feedforward in SNNs} involves multiple layers of interconnected spiking neurons. Let $N^{(l)}$ denote the number of neurons in the $l$-th layer. We represent their membrane potentials and output spikes at time step $t$ using the vector forms $\mathbf{u}^{(l)}_{[t]} \in \mathbb{R}^{N^{(l)}}$ and $\mathbf{s}^{(l)}_{[t]} \in \{0, 1\}^{N^{(l)}}$, respectively. The weights connecting layers $l-1$ and $l$ are denoted as $\mathbf{W}^{{l}}\in \mathbb{R}^{N^{(l)}\times N^{(l-1)}}$. The leaky constant of LIF neuron is given by $\lambda=1-\frac{1}{\tau}$, and the resting potential $u_{\text{rest}}=0$. Then the feedforward process, illustrated in Fig.~\ref{fig:neuron_model}~(b), is formulated as follows:
\begin{gather}
    \label{eq:feedforward}
    \mathbf{h}_{[t]}^{(l)} = 
    \underbrace{\lambda \mathbf{u}^{(l)}_{[t-1]}}_{\text{leak}} + 
    \underbrace{\mathbf{W}^{(l)} \mathbf{s}^{(l-1)}_{[t]}}_{\text{charge}}, \\
    \mathbf{s}^{(l)}_{[t]} = \underbrace{
    \Theta (
        \mathbf{h}_{[t]}^{(l)} - V_{\text{th}}
    )}_{\text{spike}}, \quad
    \mathbf{u}^{(l)}_{[t]} = 
    \mathbf{h}_{[t]}^{(l)}
    \underbrace{- V_{\text{th}}\mathbf{s}^{(l)}_{[t]}}_{\text{reset}}.
\end{gather}

\textbf{Backpropagation through time in SNNs} is shown in Fig.~\ref{fig:neuron_model}~(c). Given the gradients from the last layer $\frac{\partial \mathcal{L}}{\mathbf{s}^{(l)}_{[t]}}$, we can unfold the rolling update of membrane potential for $T$ time steps and calculate the backpropagate gradients $\frac{\partial \mathcal{L}}{\partial \mathbf{s}^{(l-1)}_{[t]}}$ and parameters gradients $\frac{\partial \mathcal{L}}{\partial \mathbf{W}^{(l)}}$ as follows\footnote{Numerator layout of matrix differentiation is used.},
\begin{equation} \footnotesize
    \frac{\partial \mathcal{L}}{\partial \mathbf{s}^{(l-1)}_{[t]}} = \sum_{k=t}^{T} 
    \underbrace{\frac{\partial \mathcal{L}}{\partial \mathbf{s}^{(l)}_{[k]}}}_{\substack{\text{gradient from} \\ \text{last layer}}}
    \underbrace{\frac{\partial \mathbf{s}^{(l)}_{[k]}}{\partial \mathbf{h}^{(l)}_{[k]}}}_{\substack{\text{surrogate} \\ \text{gradient}}}
    \Big(
        \mathbf{1} +
        \prod_{\tau=t-1}^{k-1} 
        \big( 
            \lambda - V_{\text{th}}\underbrace{\frac{\partial \mathbf{s}^{(l)}_{[\tau]}}{\partial \mathbf{h}^{(l)}_{[\tau]}}}_{\substack{\text{surrogate} \\ \text{gradient}}}
        \big)
    \Big)
    \mathbf{W}^{(l)},
\end{equation}
\begin{equation} \footnotesize
    \frac{\partial \mathcal{L}}{\partial \mathbf{W}^{(l)}} = \sum_{t=0}^T 
    \frac{\partial \mathcal{L}}{\partial \mathbf{s}^{(l)}_{[t]}}
    \frac{\partial \mathbf{s}^{(l)}_{[t]}}{\partial \mathbf{h}^{(l)}_{[t]}}
    \bigg(
        \mathbf{s}_{[t]}^{(l)} + 
        \sum_{k=0}^{t-1} 
        \Big( 
            \prod_{\tau=k}^{t-1} 
            \lambda \big( 
                1 - V_{\text{th}}
                \frac{\partial \mathbf{s}^{(l)}_{[\tau]}}{\partial \mathbf{h}^{(l)}_{[\tau]}}
            \big) 
        \Big) 
        \mathbf{s}^{(l)}_{[k]}
    \bigg).
\end{equation}
Detailed derivatives are provided in Appendix D. 

\begin{figure*}[t]
    \centering
    \includegraphics[width=\textwidth]{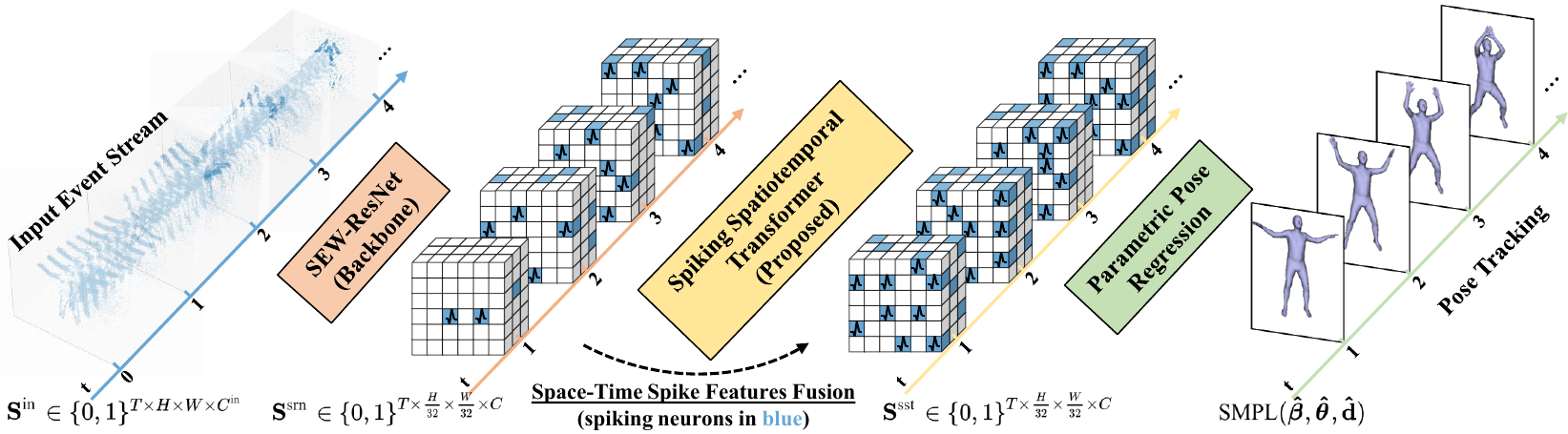}
    \caption{\textbf{Pipeline of our method}. A novel Spiking Spatiotemporal Transformer is proposed to fuse spike pose features spatiotemporally, addressing one-directional temporal dependency in SNNs.
    }
    \label{fig:pipeline}
\end{figure*}

Training SNNs from scratch is difficult mainly due to the non-differentiable property of Heaviside step function and the problem of gradient vanishing. Existing efforts summarized in~\cite{li2021differentiable} solve it by using surrogate derivatives to approximate the gradients of Heaviside step function. Following~\cite{fang2021deep}, the surrogate gradient function we used in this work is
\begin{equation}
    \frac{\partial s_{[t]}}{\partial h_{[t]}} = 
    \begin{cases}
    \frac{c}{2(1+(\frac{\pi}{2}c (h_{[t]} - V_{\text{th}}))^2)}, & \text{if } s_{[t]} = 1\\
    0,              & \text{otherwise}
    \end{cases}
\end{equation}
where $c$ is the hyper-parameter to control the smoothness of the surrogate gradients. 

\textbf{Computational and energy consumption of SNNs} are typically lower than ANNs. Following~\cite{zhou2022spikformer,yao2023attention}, the $l$-th linear layer in ANNs requires $\mathcal{O}(TN^{(l-1)} N^{(l)})$ FLOPs~\footnote{FLOPs refers to the number of floating-point operations.}, measured in multiply-and-accumulate (MAC) operations. In contrast, the $l$-th linear layer in SNNs, assuming a spiking rate of $\rho$, only requires $\mathcal{O}(\rho TN^{(l-1)} N^{(l)})$ FLOPs, measured in accumulate (AC) operations. This reduction in computational demand occurs because the computations of non-spiking neurons in the preceding layer can be skipped, \ie, the charge part $\mathbf{W}^{(l)} \mathbf{s}^{(l-1)}_{[t]}$ in Eq.~(\ref{eq:feedforward}). Regarding energy consumption, a 32-bit floating-point operation in 45nm technology consumes $E_{\text{MAC}} = 4.6pJ$ for MAC and $E_{\text{AC}} = 0.9pJ$ for AC operations~\cite{horowitz20141}.

\section{Method}
Our pipeline, illustrated in Fig.~\ref{fig:pipeline}, begins with the conversion of the input event stream into a sequence of sparse voxel grids~\cite{gallego2019event}, denoted by $\mathbf{S}^{\text{in}}\in \{0, 1\}^{T\times H\times W\times C^{\text{in}}}$. Each voxel within these grids represents a specific spatial and temporal interval. For each voxel, a value of $1$ is assigned if the number of events within it surpasses a predefined threshold; otherwise, it is set to $0$. Then SEW-ResNet~\cite{fang2021deep} is employed as the SNN backbone to extract the spike features $\mathbf{S}^{\text{srn}}$, followed by our proposed Spiking Spatiotemporal Transformer that carries bi-directional fusion of the acquired pose spike features, denoted as $\mathbf{S}^{\text{sst}}$. Finally, the spatiotemporal spike features undergo 2D average pooling and direct regression to predict the parametric 3D human poses over time. 

\subsection{Spike-Element-Wise Residual Networks}
\label{sec:sew-resnet}

SEW-ResNet~\cite{fang2021deep} serves as the backbone to extract pose spike features, denoted as $\mathbf{S}^{\text{srn}}\in \{0, 1\}^{T\times \frac{H}{32}\times \frac{W}{32}\times C}$. Across all SEW-ResNet models, the spatial dimensions of the output are reduced to $1/32$ of the input. The output channel size varies, with SEW-ResNet18 and 34 having $C=512$, while SEW-ResNet50, 101, and 152 having $C=2048$. Comprehensive architecture details of SEW-ResNet are available in Appendix A.

\begin{figure*}[t]
    \centering
    \includegraphics[width=\textwidth]{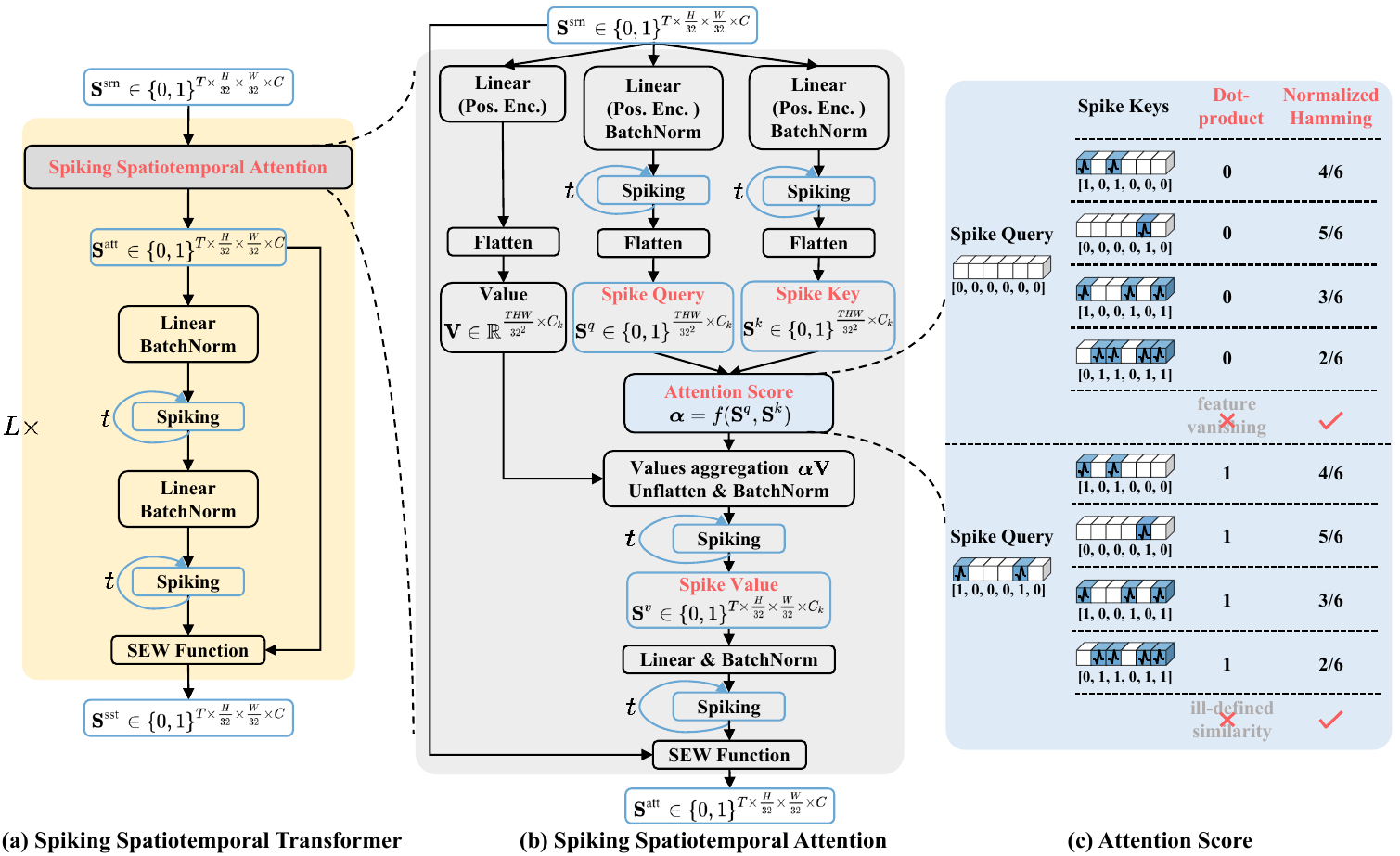}
    \caption{\textbf{Spiking Spatiotemporal Transformer}. Spiking Spatiotemporal Attention enables bi-directional flow of space-time information in SNNs. Attention Score of normalized Hamming similarity provides more accurate similarity measurement between binary spike tensors, in contrast to dot-product similarity utilized in prior works~\cite{zhou2022spikformer,zhou2023spikingformer}.}
    \label{fig:spiking-transformer}
\end{figure*}

\subsection{Spiking Spatiotemporal Transformer}
\label{sec:spike-transformer}

The architecture details are presented in Fig.~\ref{fig:spiking-transformer}~(a). Given the input spike tensor $\mathbf{S}^{\text{srn}}$, the first step is to apply spiking spatiotemporal attention to fuse bi-directional space-time features, with further details provided later. This is followed by two linear spiking layers coupled with batch normalization, commonly known as the Feed-Forward Network (FFN) in standard Transformer~\cite{vaswani2017attention}. Subsequently, the SEW Function~\cite{fang2021deep} is applied to the output of the FFN and the spiking attention output spike tensor for residual learning. Finally, we obtain the output $\mathbf{S}^{\text{sst}}\in \{0,1\}^{T\times \frac{H}{32}\times \frac{W}{32} \times C}$, which maintains the same dimensions as the input. This module can be stacked by $L$ layers, similar to~\cite{vaswani2017attention}.

\textbf{Spiking Spatiotemporal Attention} is illustrated in Fig.~\ref{fig:spiking-transformer}~(b). This module introduces self-attention that spans the entire space-time domain of spike tensors, effectively addressing the issue of one-directional temporal dependency flow in the spiking layers of SNNs. Specifically, starting with the input spike tensor $\mathbf{S}^{\text{sew}}\in \{0,1\}^{T\times \frac{H}{32}\times \frac{W}{32} \times C}$, we use two linear spiking layers followed with positional encoding and batch normalization to map the channel size from $C$ to $C_k$. Subsequently, we flatten it across the space-time dimensions to obtain the spike query and key tensors, denoted as $\mathbf{S}^{q},\mathbf{S}^{k}\in \{0,1\}^{\frac{THW}{32^2}\times C_k}$. Similarly, we obtain the real-valued tensor $\mathbf{V} \in \mathbb{R}^{\frac{THW}{32^2}\times C_k}$ by applying a non-spiking linear layer and positional encoding to transform the channel size from $C$ to $C_k$, and then flattening it across the space-time dimensions. As shown in Sec.~\ref{sec:ablation}, real Value aggregation before generating spikes presents better results. 

Next, the similarity between the spiking queries and keys is calculated using the function $\boldsymbol{\alpha}=f(\mathbf{S}^{q},\mathbf{S}^{k})$, which then serves as the attention scores for values aggregation, denoted as $\boldsymbol{\alpha}\mathbf{V}$. The details of $f(\cdot)$ will be covered later. The aggregated value tensor is then unflattened and processed through a batch normalization and spiking layer to generate the spike value tensor, denoted as $\mathbf{S}^{v}\in\{0, 1\}^{T\times \frac{H}{32}\times \frac{W}{32} \times C_k}$. Afterwards, we use a spiking linear layer with batch normalization to map the channel size from $C_k$ back to $C$. Finally, the SEW Function is applied to the attention output and the input spike tensor for residual learning, resulting in the output denoted as $\mathbf{S}^{\text{att}}\in \{0,1\}^{T\times \frac{H}{32}\times \frac{W}{32} \times C}$. Similar to~\cite{vaswani2017attention}, our model also supports multi-head attention.

Prior work~\cite{zhou2022spikformer,zhou2023spikingformer} typically generates binary spike value $\mathbf{S}^{v}$ for computational acceleration in values aggregation. Our framework also supports this architectural design. However, the ablation study in Tab.~\ref{tab:ablation-architecture} shows that using real value $\mathbf{V}$ in values aggregation significantly reduces pose tracking error, despite a slight increase in energy consumption associated with floating-point computations. Essentially, our method allows for flexibility in choosing between binary and real value $\mathbf{V}$, depending on the desired trade-off between performance and energy consumption.

\textbf{Attention scores} proposed in recent Spiking Transformers~\cite{zhou2022spikformer,zhou2023spikingformer} are based on dot-products, typically following the design of standard Transformer~\cite{vaswani2017attention}. However, dot-product similarity is not well-defined for binary spike vectors. When there are zero elements in the spike query vector, the dot-product will inevitably disregard the values of corresponding elements in the spike key vector. As an example shown in Fig.~\ref{fig:spiking-transformer}~(c), given a spike query vector, dot-product always yields the same similarity for four different spike key vectors. Notably, in the case of an all-zero spike query, the dot-product similarity score will always be zero. This means that the dot-product used in~\cite{vaswani2017attention,zhou2022spikformer} is actually unable to precisely measure the similarity between two binary spike vectors.

\begin{proposition}[Johnson–Lindenstrauss Lemma on Binary Embedding~\cite{jacques2013robust,yi2015binary}]
\label{proposition:hamming-similarity}
Define $\mathbf{q}_i, \mathbf{k}_j \in\mathbb{R}^{d_k}$ as the real-valued query and key. The corresponding binary embeddings $\mathbf{s}^q_i, \mathbf{s}^k_j \in \{0, 1\}^{C_k}$ are obtained as 
\begin{equation}
    \label{eq:mapping}
    \mathbf{s}^q_i(\mathbf{q}_i) = \text{sign}(\mathbf{Aq}_i),\quad \mathbf{s}^k_j(\mathbf{k}_j) = \text{sign}(\mathbf{Ak}_j),
\end{equation}
where $\mathbf{A}\in \mathbb{R}^{C_k \times d_k}$ is a projection matrix with each element generated independently from the normal distribution $\mathcal{N}(0, 1)$. Given that $\delta > 0$ and $\scriptstyle C_k > \frac{\log M}{\delta^2}$, we have
\begin{equation}
    g(d_{\mathcal{H}}(\mathbf{s}^q_i, \mathbf{s}^k_j) - \delta) \leq d_{\mathcal{C}}(\mathbf{q}_i, \mathbf{k}_j) \leq g(d_{\mathcal{H}}(\mathbf{s}^q_i, \mathbf{s}^k_j) + \delta),
\end{equation}
with probability at least $1-2e^{-\delta^2 C_k}$. Here $g(x)=\cos(\pi (1-x))$ is a continuous and monotone function for $x\in [0, 1]$, $M$ is the number of all possible keys and queries given by the finite training set, $d_{\mathcal{H}}\in [0, 1]$ is the normalized Hamming similarity between binary spike vectors defined as
\begin{equation}
    \label{eq:hamming-similarity}
    d_{\mathcal{H}}(\mathbf{s}_i^q, \mathbf{s}_j^k) = 1 - \frac{1}{C_k}\sum_{c=1}^{C_k} \mathds{1}(s_{ic}^q \neq s_{jc}^k),
\end{equation}
and $d_{\mathcal{C}}\in [0, 1]$ is cosine similarity between real-valued vectors defined as
\begin{equation}
    \label{eq:cosine-similarity}
    d_{\mathcal{C}}(\mathbf{q}_i, \mathbf{k}_j) =  \frac{\mathbf{q}_i^{\top}\mathbf{k}_j}{\|\mathbf{q}_i\|\|\mathbf{k}_j\|}.
\end{equation}
\end{proposition}

The proof is provided in Appendix B. 
Proposition~\ref{proposition:hamming-similarity} reveals that cosine similarity $d_{\mathcal{C}}$ between real-valued vectors is bounded within $[g(d_{\mathcal{H}} - \delta), g(d_{\mathcal{H}} + \delta)]$, where $d_{\mathcal{H}}$ is the normalized Hamming similarity between binary spike vectors. When the channel size $C_k$ is sufficiently large, $\delta$ approaches zero, implying that $d_{\mathcal{C}}$ closely approximates $g(d_{\mathcal{H}})$ with a high probability. 

Given that $g(\cdot)$ is a continuous and monotonic function, we propose a direct utilization of $d_{\mathcal{H}}$ as the score function $f(\cdot)$ in Spiking Spatiotemporal Transformer. This function guarantees accurate similarity measurement between binary spike vectors as it approximates the cosine similarity measurement between corresponding real-valued vectors via non-uniform scaling of $g(\cdot)$, behaving similarly to the scaled dot-product between real-valued vectors in standard Transformer~\cite{vaswani2017attention}. 

\textbf{The gradient of normalized Hamming similarity} does not exist since Eq.~(\ref{eq:hamming-similarity}) is a non-differentiable function. Thus we approximate it by
\begin{equation}
    \label{eq:hamming-similarity-approx}
    d_{\mathcal{H}}(\mathbf{s}_i^q, \mathbf{s}_j^k) = 1 - \frac{1}{C_k}\sum_{c=1}^{C_k} \left[ \mathbf{s}_{ic}^q\cdot(1-\mathbf{s}_{jc}^k) +  (1-\mathbf{s}_{ic}^q)\cdot\mathbf{s}_{jc}^k \right]. \nonumber
\end{equation}
As a result, the approximate gradient is given by
\begin{equation}
    \frac{\partial d_{\mathcal{H}}(\mathbf{s}_i^q, \mathbf{s}_j^k)}{\partial \mathbf{s}_i^q} = \frac{2\mathbf{s}_j^k - 1}{C_k}, \quad
    \frac{\partial d_{\mathcal{H}}(\mathbf{s}_i^q, \mathbf{s}_j^k)}{\partial \mathbf{s}_j^k} = \frac{2\mathbf{s}_i^q - 1}{C_k}. \nonumber
\end{equation}

\subsection{Parametric Pose Regression}
\label{sec:regression}
In this work, the SMPL model~\cite{loper2015smpl} is utilized as the parametric human pose and shape model. Given the input spike feature $\mathbf{S}^{\text{sst}}\in \{0,1\}^{T \times \frac{H}{32} \times \frac{W}{32}\times C}$, we apply 2D average pooling followed by three parallel linear layers to regress SMPL shape parameters $\boldsymbol{\hat \beta}\in\mathbb{R}^{T\times 10}$, SMPL pose parameters $\boldsymbol{\hat \theta}\in\mathbb{R}^{T\times 72}$ and global translations $\mathbf{\hat d}\in\mathbb{R}^{T\times 3}$ respectively. Meanwhile, we can also obtain the 3D and 2D joint positions using predefined camera parameters. 

\begin{figure}[t!]
    \centering
    \includegraphics[width=\columnwidth]{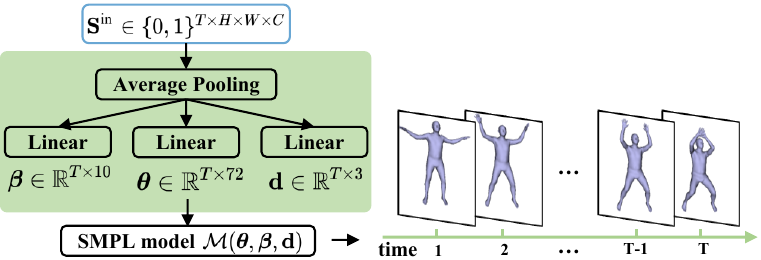}
    \caption{\textbf{Human Poses Regression.} The input spike tensor first undergoes 2D average pooling and then goes through three linear layers in parallel to regress the global translation $\mathbf{d}$ and SMPL pose and shape parameters $\boldsymbol{\theta}, \boldsymbol{\beta}$ over all $T$ time steps.}
    \label{fig:regression}
\end{figure}

The workflow is presented in Fig.~\ref{fig:regression}, where we apply 2D average pooling to the input spike tensor and then directly regress shape parameters $\boldsymbol{\hat \beta}$, pose parameters $\boldsymbol{\hat \theta}$ and global translations $\mathbf{\hat d}$ via three parallel linear layers.

\textbf{Training loss.} We use the losses of pose and shape parameters, global translations, 3D and 2D joint positions for training, which are defined as:
\begin{equation}
    \begin{aligned}
        \mathcal{L}=
        \lambda_{\text{pose}}\mathcal{L}_{\text{pose}}+
        \lambda_{\text{shape}}\mathcal{L}_{\text{shape}}+
        \lambda_{\text{trans}}\mathcal{L}_{\text{trans}}+
         \lambda_{\text{3D}}\mathcal{L}_{\text{3D}}+
        \lambda_{\text{2D}}\mathcal{L}_{\text{2D}},
        \nonumber
    \end{aligned}
\end{equation}
where $\lambda$s are the loss weights. For the poses loss, we use the 6D representation of rotations, which has been shown to perform better than the 3D axis-angle representation in~\cite{zhou2019continuity,zou2021eventhpe} for human pose estimation. Then we use the geodesic distance in $SO(3)$ to measure the difference between the predicted and target poses:
\begin{equation}
    \mathcal{L}_{\text{pose}} = \sum_{t=1}^{T}\sum_{j=1}^{24} \arccos^2\Big(\frac{\Tr\big(R^{\top}(\boldsymbol{\theta}^{j}_{[t]})R(\boldsymbol{\hat \theta}^{j}_{[t]})\big)-1}{2}\Big), \nonumber
\end{equation}
where $R(\cdot)$ is the function that transforms the 6D rotational representation to the $3\times3$ rotation matrix and $j$ is the joint index. Other losses are basically Euclidean distances between the predicted and target values as follows:
\begin{gather}
    \mathcal{L}_{\text{shape}} = \sum_{t=1}^{T} \|\boldsymbol{\beta}_{[t]} - \boldsymbol{\hat \beta}_{[t]}\|^2,\quad \mathcal{L}_{\text{trans}} = \sum_{t=1}^{T} \|\mathbf{d}_{[t]} - \mathbf{\hat d}_{[t]}\|^2, \nonumber\\
    \mathcal{L}_{\text{3D}} = \sum_{t=1}^{T} \sum_{j=1}^{24} \|\mathbf{j}^j_{\text{3D}[t]} - \mathbf{\hat j}^j_{\text{3D}[t]}\|^2, \nonumber\\
    \mathcal{L}_{\text{2D}} = \sum_{t=1}^{T} \sum_{j=1}^{24} \|\mathbf{j}^j_{\text{2D}[t]} - \mathbf{\hat j}^j_{\text{2D}[t]}\|^2. \nonumber
\end{gather}

\begin{table}[t!]
    \centering
    \caption{\textbf{Comparative Analysis of Event-Based 3D Human Pose Datasets}: real or synthetic data (R/S), number of subjects (Sub \#), number of event streams (Str \#), total time length of all event streams in hours (Len), average time length in minutes (AvgLen), SMPL annotations (SMPL Pose).}
    \setlength{\tabcolsep}{1mm}
    \renewcommand{\arraystretch}{1.2}
    \resizebox{0.5\textwidth}{!}{
    \begin{tabular}{@{}|c|cccccc|@{}}
    \bottomrule \hline
        \makecell[c]{Dataset} & \makecell[c]{R/S} & \makecell[c]{Sub \\ \#} & \makecell[c]{Str \\ \#} & \makecell[c]{Len\\(hrs)} & \makecell[c]{AvgLen\\(mins)} & \makecell[c]{SMPL\\Pose} \\
    \hline
        \makecell[c]{DHP19~\cite{calabrese2019dhp19}} & \makecell[c]{Real} & \makecell[c]{17} & \makecell[c]{33} & \makecell[c]{0.80} & \makecell[c]{1.46}  & \makecell[c]{\newcrossmark} \\
        \makecell[c]{MMHPSD~\cite{zou2021eventhpe}} & \makecell[c]{Real} & \makecell[c]{15} & \makecell[c]{178} & \makecell[c]{4.39} & \makecell[c]{1.48}  & \makecell[c]{\newcheckmark} \\
    \hline
        \makecell[c]{SynEventHPD (Ours)} & \makecell[c]{Syn} & \makecell[c]{47} & \makecell[c]{9197} & \makecell[c]{45.72} & \makecell[c]{0.30} & \makecell[c]{\newcheckmark} \\
        \makecell[l]{\ \ - 1) EventH36M} & \makecell[c]{Syn} & \makecell[c]{7} & \makecell[c]{835} & \makecell[c]{12.46} & \makecell[c]{0.90} & \makecell[c]{\newcheckmark} \\
        \makecell[l]{\ \ - 2) EventAMASS} & \makecell[c]{Syn} & \makecell[c]{13} & \makecell[c]{8028} & \makecell[c]{23.54} & \makecell[c]{0.18}  & \makecell[c]{\newcheckmark} \\
        \makecell[l]{\ \ - 3) EventPHSPD} & \makecell[c]{Syn} & \makecell[c]{12} & \makecell[c]{156} & \makecell[c]{5.33} & \makecell[c]{2.05}  & \makecell[c]{\newcheckmark} \\
        \makecell[l]{\ \ - 4) SynMMHPSD} & \makecell[c]{Syn} & \makecell[c]{15} & \makecell[c]{178} & \makecell[c]{4.39} & \makecell[c]{1.48}  & \makecell[c]{\newcheckmark} \\
    \hline \toprule  
    \end{tabular}
    }
    \label{tab:dataset-summary}
\end{table}

\subsection{Our SynEventHPD Dataset}
\begin{figure}[t!]
    \centering
    \includegraphics[width=0.5\textwidth]{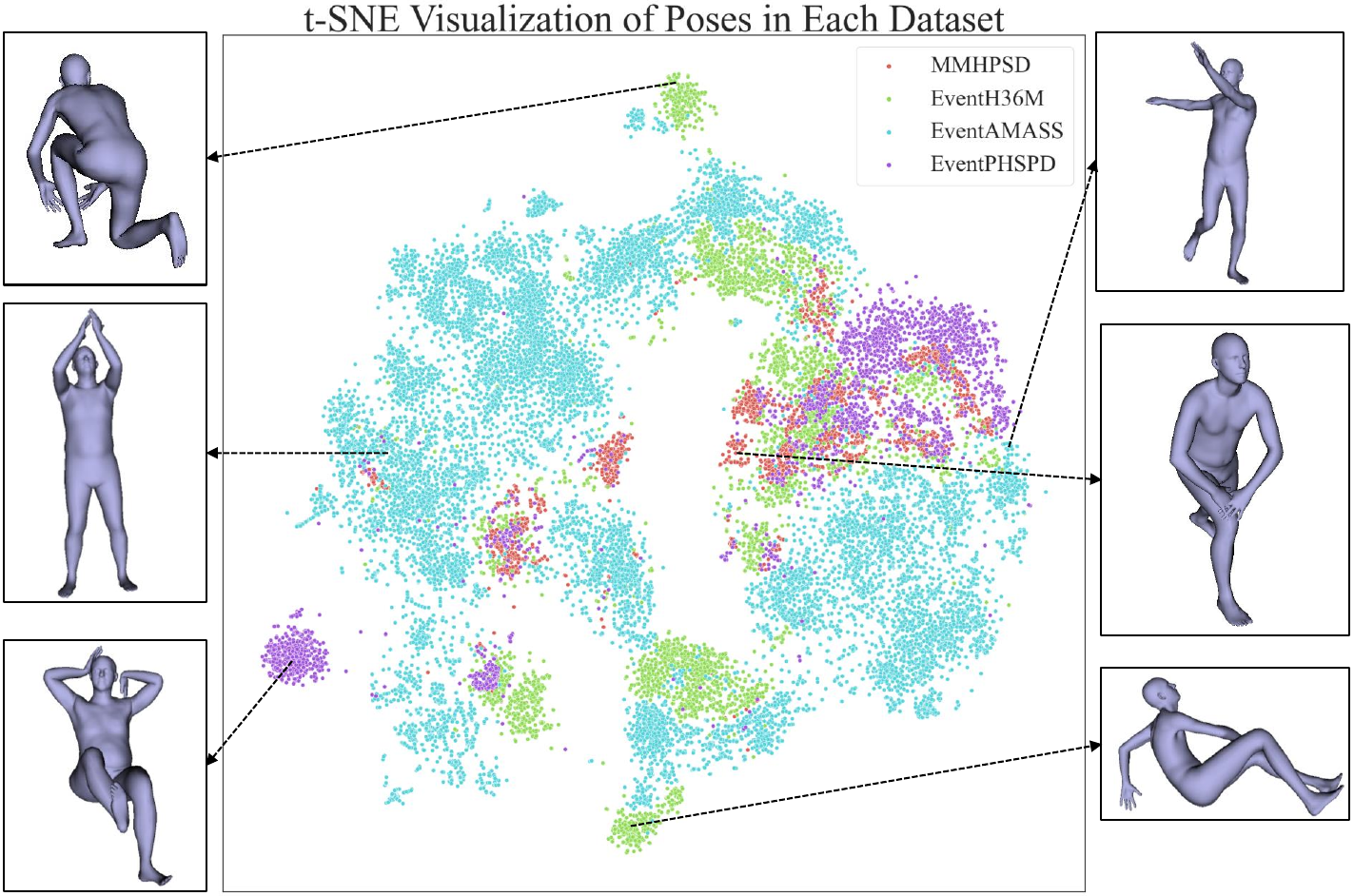}
    \caption{\textbf{t-SNE Visualization of Poses} from each sub-dataset in our SynEventHPD dataset. MMHPSD~\cite{gehrig2020video} only covers a small area, while our SynEventHPD dataset, by including 4 sub-datasets (EventH36M, EventAMASS, EventPHSPD and SynMMHPSD), contains a wide range of poses. This highlights the rich variety of poses provided in SynEventHPD.}
    \label{fig:poses-tsne}
\end{figure}

Currently, the largest event-based dataset for human pose estimation is MMHPSD~\cite{zou2021eventhpe}, which includes 15 subjects, 21 different actions and a total of 4.39 hours of event streams. However, the limited variety of motions in this dataset restricts the generalization ability of trained models. To address this issue, we propose to synthesize events from multiple motion capture datasets, including \textit{Human3.6M~\cite{ionescu2013human3}, AMASS~\cite{mahmood2019amass}, PHSPD~\cite{zou2022human} and MMHPSD~\cite{zou2021eventhpe}}, to construct a large-scale synthetic dataset. The RGB or gray-scale videos from Human3.6M, PHSPD and MMHPSD datasets are directly used for events synthesis. Since AMASS dataset only provides motion capture data without corresponding videos, we animate one of 13 avatars for each motion sequence and render corresponding RGB video before applying it to events synthesis. Our synthetic dataset, called SynEventHPD, is a meta dataset consisting of 4 sub-datasets: EventH36M, EventAMASS, EventPHSPD and SynMMHPSD. In total, it contains 45.72 hours of event streams, more than 10 times of MMHPSD. The detailed synthesizing process is described in Appendix C. 
More details are summarized in Tab.~\ref{tab:dataset-summary}. The distribution of poses across all these datasets are visualized in Fig.~\ref{fig:poses-tsne} to highlight the variety. 

\section{Experiments}
\subsection{Empirical Results on Real MMHPSD Dataset}
\label{sec:exp}

\textbf{Implementation Details.} For a fair comparison with prior works~\cite{xu2020eventcap,zou2021eventhpe}, we follow the train and test set split for MMHPSD dataset from~\cite{zou2021eventhpe}. We present the results of model trained with $T=8$ and $T=64$ time steps, which stands for 2 and 8 seconds of input event stream respectively. For event stream preprocessing, we convert events into $T$ voxel grids of size $256\times 256\times 4$. Empirically, we find that $4$ is the best choice, as higher values do not show performance improvements. We use parametric LIF neuron with soft reset and retain the backpropagation of reset path in SNNs, where SpikingJelly~\cite{SpikingJelly} is used to implement the model. To fairly compare with other baselines in terms of the number of parameters, we use SEW-ResNet50~\cite{fang2021deep} as the backbone and configure the Spiking Spatiotemporal Transformer with hidden dimension $C_k$ set to be 1024, along with 1 attention head and 2 layers, resulting in 47.7M parameters.

\begin{table}[t]
    \centering
    \caption{\textbf{Architecture of Different Models.}}
    \setlength{\tabcolsep}{0.5mm}
    \renewcommand{\arraystretch}{2}
    \resizebox{0.5\textwidth}{!}{
    \begin{tabular}{@{}|c|c|c|@{}}
    \bottomrule \hline
        Model & \makecell[c]{ANN/\\SNN} & \makecell[c]{Architecture}\\
    \hline
        VIBE~\cite{kocabas2019vibe} & \multirow{6}{*}{ANN} & \makecell[c]{\texttt{ResNet50 + GRU-1024hidden-bidirectional}}\\
    \cline{1-1}\cline{3-3}
        MPS~\cite{wei2022capturing} & & \makecell[c]{\texttt{ResNet50 + Temporal Attentive Module}}\\
    \cline{1-1}\cline{3-3}
        EventHPE~\cite{zou2021eventhpe} & & \makecell[c]{\texttt{ResNet50 + GRU-1024hidden-bidirectional}}\\
    \cline{1-1}\cline{3-3}
        ResNet-GRU~\cite{kocabas2019vibe} & & \makecell[c]{\texttt{ResNet50 + GRU-1layer-1024hidden-bidirectional}}\\
    \cline{1-1}\cline{3-3}
        ResNet-TF~\cite{carion2020end} & & \makecell[c]{\texttt{ResNet50 +}\\ \texttt{Transformer-Encoder-2layers-384hidden-8heads +}\\ \texttt{Transformer-Decoder-2layers-384hidden-8heads}}\\
    \cline{1-1}\cline{3-3}
        Video-Swin~\cite{liu2021swin} & & \makecell[c]{\texttt{Swin\_Tiny-96hidden-depths=[2,2,6,2]-}\\\texttt{heads=[3,6,12,24]-window\_size=(8,8,8)}}\\
    \hline
        SEW-ResNet-TF & \makecell[c]{Mix} & \makecell[c]{\texttt{SEW-ResNet50 +}\\ \texttt{Transformer-Encoder-2layers-384hidden-8heads +}\\ \texttt{Transformer-Decoder-2layers-384hidden-8heads}}\\
    \hline
        MA-SNN~\cite{yao2023attention} & \multirow{4}{*}{SNN} & \makecell[c]{\texttt{SEW-ResNet50 + Multi-dimensional Attention}}\\
    \cline{1-1}\cline{3-3}
        Spikformer~\cite{zhou2022spikformer} & & \makecell[c]{\texttt{16x16patch-6layers-1024hidden}}\\
    \cline{1-1}\cline{3-3}
        Spikingformer~\cite{zhou2023spikingformer} & & \makecell[c]{\texttt{16x16patch-6layers-1024hidden}}\\
    \cline{1-1}\cline{3-3}
        Ours & & \makecell[c]{\texttt{ResNet50 +  SpikingSpatiotemporal}\\\texttt{Transformer-2layers-1024-1head}}\\
    \hline \toprule 
    \end{tabular}}
    \label{tab:architecture}
\end{table}

\begin{table*}[h!]
    \centering
    \caption{\textbf{Quantitative results of human pose tracking on MMHPSD dataset.} The input includes video (\textit{V}), first gray-scale frame (\textit{G}), events (\textit{E}) and their combination (\textit{V+E}, \textit{G+E}). 
    We present two video-based ANN models~\cite{kocabas2019vibe,wei2022capturing,shen2023global} and two SOTA ANN-based methods for event-based pose tracking~\cite{xu2020eventcap,zou2021eventhpe}. 
    Additionally, we include three ANN models~\cite{kocabas2019vibe,carion2020end,liu2021swin}, one Mix model of ANNs and SNNs (\textit{SEW-ResNet-TF}) and five SNN models~\cite{rueckauer2017conversion,fang2021deep,yao2023attention,zhou2022spikformer,zhou2023spikingformer} as benchmarks for events-only pose tracking task. 
    \underline{Underline} denotes the best results, except for $\boldsymbol{^{\dagger}}$\textit{EventHPE(GT)} that means the ground-truth starting pose is known in EventHPE, serving as the upper bound among all the methods.}
    \setlength{\tabcolsep}{0.5mm}
    \renewcommand{\arraystretch}{1.2}
    \resizebox{\textwidth}{!}{
    \begin{tabular}{@{}|c|c|c|c|ccccc|ccccc|@{}}
    \bottomrule \hline 
        \multirow{2}{*}{Method} & \multirow{2}{*}{\makecell[c]{ANN/\\SNN}} & \multirow{2}{*}{Input} & \multirow{2}{*}{Params} & \multicolumn{5}{c|}{T=8 (1 sec)} & \multicolumn{5}{c|}{T=64 (8 secs)}\\
        \cline{5-14}
        & & & & \makecell[c]{FLOPs} & \makecell[c]{Engy} & \makecell[c]{MPJPE $\downarrow$} & \makecell[c]{PEL-MPJPE $\downarrow$} & \makecell[c]{PA-MPJPE $\downarrow$} & \makecell[c]{FLOPs} & \makecell[c]{Engy} & \makecell[c]{MPJPE $\downarrow$} & \makecell[c]{PEL-MPJPE $\downarrow$} & \makecell[c]{PA-MPJPE $\downarrow$}\\
    \hline
        VIBE~\cite{kocabas2019vibe} & \multirow{3}{*}{ANN} & \makecell[c]{V} & \makecell[c]{48.3M} & \makecell[c]{43.4G} & \makecell[c]{0.19} & \makecell[c]{-} & \makecell[c]{73.1} & \makecell[c]{50.9} & \makecell[c]{344.9G} & \makecell[c]{1.58} & \makecell[c]{-} & \makecell[c]{75.4} & \makecell[c]{53.6}\\
        MPS~\cite{wei2022capturing} & & \makecell[c]{V} & \makecell[c]{39.6M} & \makecell[c]{45.6G} & \makecell[c]{0.20} & \makecell[c]{-} & \makecell[c]{68.0} & \makecell[c]{48.2} & \makecell[c]{348.3G} & \makecell[c]{1.60} & \makecell[c]{-} & \makecell[c]{69.2} & \makecell[c]{50.1}\\
        GLoT~\cite{shen2023global} & & \makecell[c]{V} & \makecell[c]{40.5M} & \makecell[c]{71.3G} & \makecell[c]{0.48} & \makecell[c]{-} & \makecell[c]{65.2} & \makecell[c]{46.3} & \makecell[c]{660.3G} & \makecell[c]{3.04} & \makecell[c]{-} & \makecell[c]{63.2} & \makecell[c]{46.1}\\
    \hline
        EventCap(VIBE)~\cite{xu2020eventcap} & \multirow{5}{*}{ANN} & \makecell[c]{V+E} & \makecell[c]{48.3M} & \makecell[c]{185.0G} & \makecell[c]{0.85} & \makecell[c]{-} & \makecell[c]{71.9} & \makecell[c]{50.4} & \makecell[c]{1477.7G}  & \makecell[c]{6.79} & \makecell[c]{-} & \makecell[c]{74.1} & \makecell[c]{52.9}\\
        EventCap(MPS)~\cite{xu2020eventcap} & & \makecell[c]{V+E} & \makecell[c]{39.6M} & \makecell[c]{187.2G} & \makecell[c]{0.86} & \makecell[c]{-} & \makecell[c]{66.6} & \makecell[c]{47.8} & \makecell[c]{1481.1G}  & \makecell[c]{6.81} & \makecell[c]{-} & \makecell[c]{68.1} & \makecell[c]{49.5}\\
        EventHPE(VIBE)~\cite{zou2021eventhpe} & & \makecell[c]{G+E} & \makecell[c]{49.0M} & \makecell[c]{49.0G} & \makecell[c]{0.22} & \makecell[c]{-} & \makecell[c]{69.6} & \makecell[c]{48.9} & \makecell[c]{354.0G} & \makecell[c]{1.62} & \makecell[c]{-} & \makecell[c]{71.6} & \makecell[c]{50.2}\\
        EventHPE(MPS)~\cite{zou2021eventhpe} & & \makecell[c]{G+E} & \makecell[c]{39.6M} & \makecell[c]{49.3G}  & \makecell[c]{0.22} & \makecell[c]{-} & \makecell[c]{65.1} & \makecell[c]{46.5} & \makecell[c]{354.2G} & \makecell[c]{1.63} & \makecell[c]{-} & \makecell[c]{66.8} & \makecell[c]{48.1}\\
        $\boldsymbol{^{\dagger}}$EventHPE(GT)~\cite{zou2021eventhpe} & & \makecell[c]{G+E} & \makecell[c]{46.9M} & \makecell[c]{-} & \makecell[c]{-} & \makecell[c]{71.8} & \makecell[c]{55.0} & \makecell[c]{43.9} & \makecell[c]{-} & \makecell[c]{-} & \makecell[c]{74.5} & \makecell[c]{58.1} & \makecell[c]{45.3}\\
    \hline
        ResNet-GRU~\cite{kocabas2019vibe} & \multirow{3}{*}{ANN}  & \makecell[c]{E} & \makecell[c]{46.9M} & \makecell[c]{43.6G}  & \makecell[c]{0.20} & \makecell[c]{111.2} & \makecell[c]{60.0} & \makecell[c]{45.3} & \makecell[c]{348.6G}  & \makecell[c]{1.60} & \makecell[c]{115.0} & \makecell[c]{64.2} & \makecell[c]{49.5}\\
        ResNet-TF~\cite{carion2020end} & & \makecell[c]{E} & \makecell[c]{41.3M} & \makecell[c]{50.5G}  & \makecell[c]{0.23} & \makecell[c]{108.5} & \makecell[c]{59.9} & \makecell[c]{\underline{\textbf{44.1}}} & \makecell[c]{403.8G}  & \makecell[c]{1.85} & \makecell[c]{114.2} & \makecell[c]{66.0} & \makecell[c]{50.1}\\
        Video-Swin~\cite{liu2021swin} & & \makecell[c]{E} & \makecell[c]{48.9M} & \makecell[c]{44.7G}  & \makecell[c]{0.20} & \makecell[c]{124.1} & \makecell[c]{66.5} & \makecell[c]{49.0} & \makecell[c]{359.6G} & \makecell[c]{1.65} & \makecell[c]{130.9} & \makecell[c]{72.5} & \makecell[c]{53.1}\\
    \hline
        SEW-ResNet-TF & \makecell[c]{Mix} & \makecell[c]{E} & \makecell[c]{47.0M} & \makecell[c]{24.5G}  & \makecell[c]{0.097} & \makecell[c]{110.8} & \makecell[c]{58.9} & \makecell[c]{44.2} & \makecell[c]{199.7G} & \makecell[c]{0.79} & \makecell[c]{113.2} & \makecell[c]{65.3} & \makecell[c]{49.3}\\
    \hline
        ANN2SNN~\cite{rueckauer2017conversion} & \multirow{5}{*}{SNN} & \makecell[c]{E} & \makecell[c]{46.9M} & \makecell[c]{12.5G}  & \makecell[c]{0.011} & \makecell[c]{140.3} & \makecell[c]{74.1} & \makecell[c]{55.8} & \makecell[c]{98.8G} & \makecell[c]{0.089} & \makecell[c]{148.2} & \makecell[c]{81.1} & \makecell[c]{60.9}\\
        SEW-ResNet~\cite{fang2021deep} & & \makecell[c]{E} & \makecell[c]{\underline{\textbf{25.8M}}} & \makecell[c]{9.1G}  & \makecell[c]{0.0082} & \makecell[c]{116.8} & \makecell[c]{63.2} & \makecell[c]{49.1} & \makecell[c]{\underline{\textbf{56.7G}}} & \makecell[c]{\underline{\textbf{0.051}}} &\makecell[c]{125.8} & \makecell[c]{68.4} & \makecell[c]{52.1}\\
        MA-SNN~\cite{yao2023attention} & & \makecell[c]{E} & \makecell[c]{30.2M} & \makecell[c]{\underline{\textbf{8.3G}}}  & \makecell[c]{\underline{\textbf{0.0080}}} & \makecell[c]{115.2} & \makecell[c]{62.7} & \makecell[c]{48.9} & \makecell[c]{59.3G} & \makecell[c]{0.057} & \makecell[c]{122.1} & \makecell[c]{66.7} & \makecell[c]{50.7}\\
        Spikformer~\cite{zhou2022spikformer} & & \makecell[c]{E} & \makecell[c]{36.8M} & \makecell[c]{13.2G}  & \makecell[c]{0.012} & \makecell[c]{114.3} & \makecell[c]{62.5} & \makecell[c]{48.6} & \makecell[c]{96.3G} & \makecell[c]{0.086} & \makecell[c]{121.2} & \makecell[c]{66.2} & \makecell[c]{50.0}\\
        Spikingformer~\cite{zhou2023spikingformer} & & \makecell[c]{E} & \makecell[c]{37.1M} & \makecell[c]{12.5G}  & \makecell[c]{0.012} & \makecell[c]{113.6} & \makecell[c]{62.0} & \makecell[c]{48.2} & \makecell[c]{86.4G} & \makecell[c]{0.078} & \makecell[c]{119.5} & \makecell[c]{65.4} & \makecell[c]{49.5}\\
    \hline
        Ours & \makecell[c]{SNN} & \makecell[c]{E} & \makecell[c]{47.7M} & \makecell[c]{9.4G}  & \makecell[c]{0.0083} & \makecell[c]{\underline{\textbf{107.2}}} & \makecell[c]{\underline{\textbf{58.7}}} & \makecell[c]{\underline{\textbf{44.1}}} & \makecell[c]{63.4G} & \makecell[c]{0.058} & \makecell[c]{\underline{\textbf{111.7}}} & \makecell[c]{\underline{\textbf{61.9}}} & \makecell[c]{\underline{\textbf{45.7}}}\\
    \hline \toprule 
    \end{tabular}
    }
    \label{tab:pose-estimation}
\end{table*}

During training, to ensure robustness against both fast and slow motions, we augment the training samples in two ways: randomly selecting event stream of (0.5, 1, 2, 3) seconds for $T=8$ and (4, 8, 16, 32) seconds for $T=64$ as the input, spatially rotating the voxel grid with a random degree between -20 and 20. We train the $T=8$ and $T=64$ models for 20 and 25 epochs respectively with batch size being 8 on a single NVIDIA A100 80GB GPU. The learning rate starts from 0.01 and is scheduled by CosineAnnealingLR, with maximum epoch of 21 and 26. The loss weights $\lambda_{\text{pose}}$, $\lambda_{\text{shape}}$, $\lambda_{\text{trans}}$, $\lambda_{\text{3D}}$ and $\lambda_{\text{2D}}$ are set to be 10, 1, 50, 1 and 10 respectively. We adjust the weights so that the individual loss components have similar magnitudes, preventing any single loss term from dominating the optimization process. For testing, 1 and 8-second event streams are used for $T=8$ and $T=64$ models respectively.

We will compare our SNN model with a few popular ANN and SNN models as benchmarks. The detailed architecture of these models is presented in Tab.~\ref{tab:architecture}. There are three types of models for comparison in our work: (1) ANN means the entire architecture is built on ANNs, including ResNet, GRU, Vanilla Transformer and Video-Swin Transformer. (2) Mix means the encoder is based on SNNs while the decoder remains ANN-based. (3) SNN means the entire architecture relies solely on SNNs, such as MA-SNN, Spikformer and our method. The settings for training the ANN models mostly follow those of our approach, except the learning rate, which starts from 0.0001 and is scheduled by StepLR with a 0.1 decay after 15 and 20 epochs for both T=8 and T=64, respectively. This is because ANN models do not converge well using a higher learning rate of 0.001. The models are trained on a single NVIDIA A100 80GB GPU. The settings for training the SNN models follow those of our approach for fair comparison.

\textbf{Evaluation metrics.} Following~\cite{kocabas2019vibe,zou2021eventhpe}, we report Mean Per Joint Position Error (MPJPE), PELvis-aligned MPJPE (PEL-MPJPE) and Procrustes-Aligned MPJPE (PA-MPJPE) for evaluation. PA-MPJPE compares predicted and target pose after rotation and translation alignment, while PEL-MPJPE compares after only translation alignment of root joint. We also present MAC or AC (FLOPs) count and energy consumption (Engy) in joules ($J$) to demonstrate the efficiency of SNNs.

\begin{figure*}[h!]
    \centering
    \includegraphics[width=\textwidth]{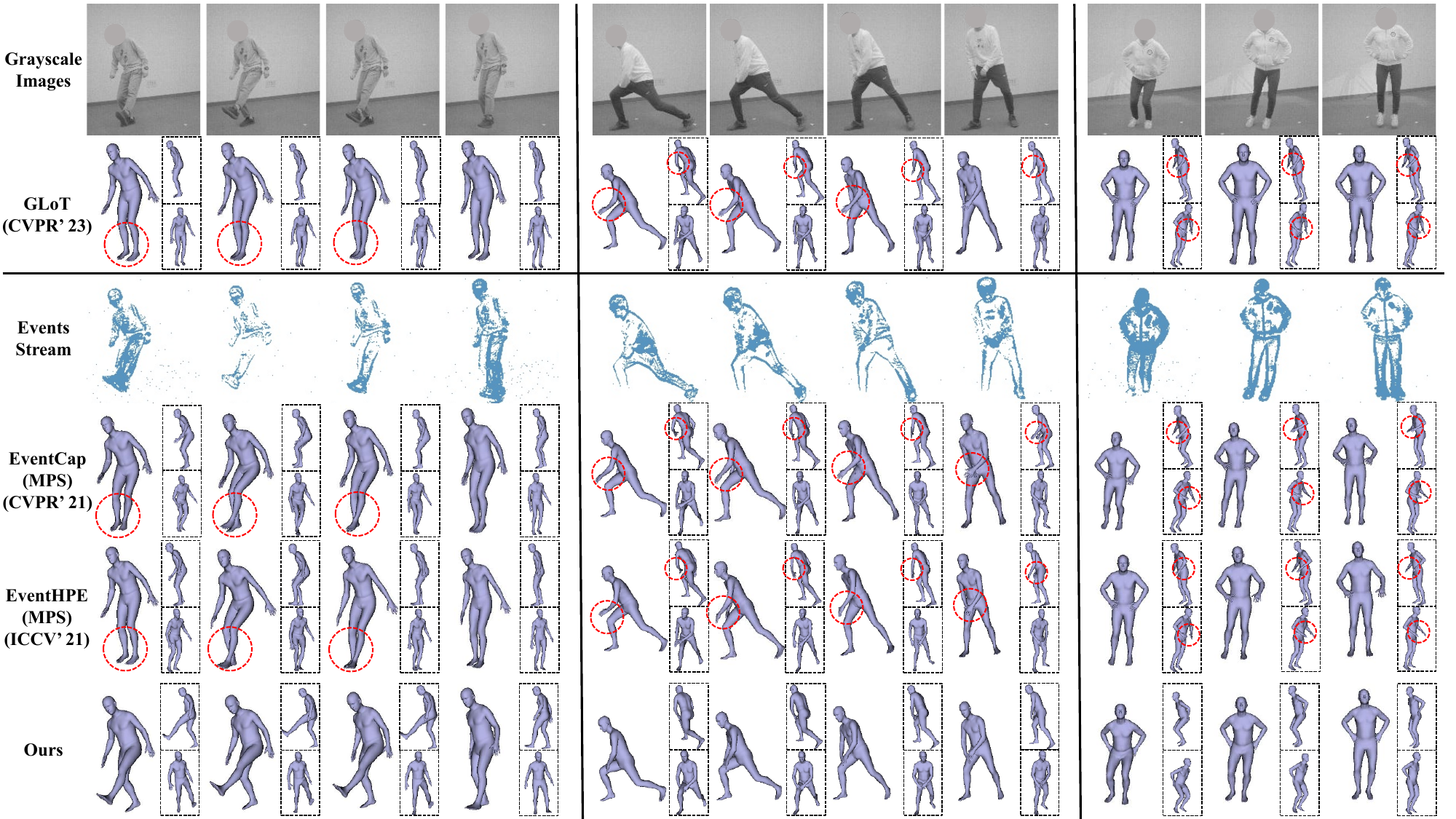}
    \caption{\textbf{Qualitative results}. Two side views are shown in dashed boxes. Prior works, EventCap and EventHPE, tend to suffer from the inaccurate poses provided by pre-trained ANN method MPS, resulting in sub-optimal pose tracking outcomes.}
    \label{fig:vis-pose}
\end{figure*}

\textbf{Comparison with SOTA methods.} 
We compare our method with four prior works to highlight the competency of solely using event signals for human pose tracking: VIBE~\cite{kocabas2019vibe}, MPS~\cite{wei2022capturing}, GLoT~\cite{shen2023global}, EventCap~\cite{xu2020eventcap}, and EventHPE~\cite{zou2021eventhpe}. 
In Tab.~\ref{tab:pose-estimation}, we use \textit{V}, \textit{G} and \textit{E} to represent the input data of gray-scale video, first gray-scale frame and event streams respectively. \textit{VIBE}, \textit{MPS} and \textit{GLoT} are applied as the most recent video-based baselines with ResNet50 as the backbone. Note that both methods use weak camera model without global translation, so we will not report their MPJPE. 
For EventCap, initial poses are extracted from the gray-scale video using pre-trained VIBE and MPS methods, denoted as \textit{EventCap(VIBE)} and \textit{EventCap(MPS)}. Notably, EventCap's iterative optimization approach typically requires higher FLOPs compared to end-to-end methods, as indicated in Table~\ref{tab:pose-estimation}.
Similarly, for EventHPE, we utilize VIBE and MPS for the starting pose extraction, denoted as \textit{EventHPE(VIBE)} and \textit{EventHPE(MPS)}. Additionally, we present results for EventHPE with ground-truth starting pose known, denoted as \textit{EventHPE(GT)}, which serves as an upper bound assumption due to its perfect information of the starting pose.

As shown in Tab.~\ref{tab:pose-estimation}, when comparing the latest MPS with VIBE, EventCap(MPS) with EventCap(VIBE), and EventHPE(MPS) with EventHPE(VIBE), we can observe that  the former consistently outperforms the latter by a similar margin for both $T=8$ and $T=64$, indicating that SOTA methods~\cite{xu2020eventcap,zou2021eventhpe} are significantly impacted by the accuracy of initial poses extracted by the pre-trained methods. Inaccurate initial poses may trap the following optimization process in local minima, yielding only marginal improvements. This is further illustrated in Fig.~\ref{fig:vis-pose}, where the inaccurate initial poses by MPS lead to sub-optimal pose tracking outcomes. 

In constrast, our SNNs model achieves the best performance with the smallest gap to the upper bound EventHPE(GT), while using only a small fraction of FLOPs and energy required by these SOTA ANN-based methods~\cite{xu2020eventcap,zou2021eventhpe}. The key advantage of our method lies in leveraging the inherent sparsity of event signals. Specifically, dense images require 8 bits per pixel, resulting in 1.5M bits for an input image with a resolution of $256\times 256\times 3$. In contrast, sparse event data requires only 1 bit per pixel (i.e., 0 or 1), leading to a significantly lower memory footprint of 0.26M bits for an event volume of $256\times 256\times 4$. Notably, pixels with a value of 0 indicate the absence of events, allowing computations over these pixels to be skipped in SNNs, further enhancing efficiency. However, existing ANN-based approaches overlook this inherent sparsity, limiting their ability to develop a highly efficient pose tracking framework.

Based on the above analysis, the latency varies when comparing our method with SOTA approaches due to two key factors. First, transmission latency is significantly reduced in our event-based framework, as dense images require 1.5M bits for transmission, while event data requires only 0.26M bits, leading to a $5.7\times$ lower transmission latency. Second, processing latency is also lower in our SNN model, which requires only 9.4G FLOPs for an 8-frame input compared to 49.3G FLOPs in the baseline EventHPE (MPS) method, resulting in a $5.2\times $ reduction in processing latency due to the sparse representation of spiking neurons. Additionally, this latency can be further minimized when deployed on specialized neuromorphic hardware~\cite{pei2019towards}.

\textbf{Comparison with ANN models.}
To further illustrate the advantages of SNNs over ANNs in events-only human pose tracking, we compare our model with three popular ANN models. The first, \textit{ResNet-GRU}, combines a spatial ResNet with a temporal GRU, as used in~\cite{kocabas2019vibe, zou2021eventhpe}, leveraging recurrent modeling to capture sequential patterns. The second, \textit{ResNet-TF}, integrates a spatial ResNet with a standard Transformer~\cite{vaswani2017attention} for temporal modeling, as seen in DETR~\cite{carion2020end}, utilizing self-attention to capture long-range dependencies. The third, \textit{Video-Swin}, is the Video Swin Transformer~\cite{liu2021swin}, which applies shift-window attention for spatiotemporal feature fusion. For a fair comparison, we select the architecture with around 45M parameters for all the models.

From Table~\ref{tab:pose-estimation}, our method outperforms ResNet-GRU and Video-Swin with slightly lower pose errors at $T=8$, while competing closely with ResNet-TF, achieving 44.1mm in PA-MPJPE. However, our model consumes only about 20\% of the FLOPs and 4\% of the energy required by ResNet-TF. For $T=64$, where longer temporal dependencies are crucial, the performance decline of ANN models is notably larger than our SNN model, with a drop of over 4.1mm vs. 1.5mm in PA-MPJPE. These results demonstrate our approach's efficiency in encoding long-term temporal dependencies within event streams.

\textbf{Comparison with SNN models.}
We compare our model with six recent SNN model to highlight its superiority in human pose tracking. \textit{SEW-ResNet-TF} serves as a hybrid baseline, combining SEW-ResNet as the SNN backbone with an ANN-based standard Transformer, leveraging self-attention for temporal modeling. \textit{ANN2SNN} converts a trained ANN model, ResNet-GRU, into an SNN using the method from~\cite{rueckauer2017conversion}, thereby inheriting the GRU’s recurrent temporal modeling capabilities. \textit{SEW-ResNet}~\cite{fang2021deep}, the backbone of our approach without the Spiking Spatiotemporal Transformer, processes temporal information implicitly through temporal spiking mechanism. \textit{MA-SNN}~\cite{yao2023attention} employs multi-dimensional attention within an SNN framework, using SEW-ResNet50 for a fair comparison. \textit{Spikformer}~\cite{zhou2022spikformer} introduces an SNN vision transformer (ViT)~\cite{dosovitskiy2020image}, incorporating spike-based self-attention for sequence modeling, while \textit{Spikingformer}~\cite{zhou2023spikingformer} further enhances residual learning to improve temporal feature extraction. Both transformer-based models adopt dot-product similarity directly in their self-attention mechanisms to model long-range dependencies.

From Tab.\ref{tab:pose-estimation}, compared to SEW-ResNet-TF, our approach shows slightly lower pose errors with less than 50\% of the FLOPs. However, ANN2SNN, while excelling in image classification, performs poorly in fine-grained task of pose tracking due to the quantization errors during conversion. Compared with SEW-ResNet, our method achieves significantly lower pose errors, highlighting the importance of bi-directional space-time information fusion by our approach. Although MA-SNN requires fewer FLOPs than our approach due to its lower spiking rate of 16.4\% vs. ours of 22.6\%, its performance is still inferior. Moreover, our method exhibits lower pose errors than Spikformer and Spikingformer, attributed to our normalized Hamming similarity in the spiking attention module vs. the ill-posed dot-product similarity.

\subsection{Empirical Results on Synthetic SynEventHPD Dataset}
\label{sec:compare-train-synthetic}

Although this dataset covers a variety of motions, as illustrated in Fig.~\ref{fig:poses-tsne}, the potential domain gap between synthetic and real events data remains an open question. In this section, we aim to demonstrate the value of our SynEventHPD dataset. We select five models as baselines, including one ANNs model (\textit{ResNet-GRU}~\cite{kocabas2019vibe}), one mixed model (\textit{SEW-ResNet-TF}) and three SNNs models (\textit{SEW-ResNet}~\cite{fang2021deep} and \textit{Ours}). All models are evaluated on the real MMHPSD test set, but trained using either the real MMHPSD train set (\textit{Real}), the synthetic SynEventHPD dataset (\textit{Syn}) or a combination of both synthetic dataset and the real train set (\textit{Real\&Syn}). Furthermore, to further illustrate the effectiveness of our proposed dataset, we employ the domain adaptation (DA)~\cite{sun2023domain} to mitigate the potential domain gap that may arise during the training of hybrid datasets.

\begin{table}[t]
    \centering
    \caption{\textbf{Effectiveness of our SynEventHPD dataset.} The values in the bracket are the improvements of each model trained with \textit{Syn\&Real} compared to trained with \textit{Real} only. \textit{DA} means domain adaptation~\cite{sun2023domain} applied during training.}
    \setlength{\tabcolsep}{0.2mm}
    \renewcommand{\arraystretch}{1.2}
    \resizebox{\columnwidth}{!}{
    \begin{tabular}{@{}|c|c|c|ccc|@{}}
    \bottomrule \hline
        \multirow{2}{*}{Model} & \multirow{2}{*}{\makecell[c]{ANN/\\SNN}} & \multirow{2}{*}{\makecell[c]{Training\\Set}} &  \multicolumn{3}{c|}{T=8 (1 sec)}\\
        \cline{4-6}
        & & & \makecell[c]{MPJPE $\downarrow$} & \makecell[c]{PEL-MPJPE $\downarrow$} & \makecell[c]{PA-MPJPE $\downarrow$}\\
    \hline
        \multirow{3}{*}{\makecell[c]{ResNet-GRU~\cite{kocabas2019vibe}}} & \multirow{3}{*}{\makecell[c]{ANN}} & \makecell[c]{Syn} & \makecell[c]{113.6} & \makecell[c]{62.2} & \makecell[c]{47.5} \\
        & & \makecell[c]{Real} & \makecell[c]{111.2} & \makecell[c]{60.0} & \makecell[c]{45.3} \\
        & & \makecell[c]{Real\&Syn} & \makecell[c]{105.4 (5.8)} & \makecell[c]{58.9 (1.1)} & \makecell[c]{44.6 (0.7)} \\
    \hline
        \multirow{3}{*}{\makecell[c]{SEW-ResNet-TF}} & \multirow{3}{*}{\makecell[c]{Mix}} & \makecell[c]{Syn} & \makecell[c]{114.1} & \makecell[c]{60.6} & \makecell[c]{45.5} \\
        & & \makecell[c]{Real} & \makecell[c]{110.8} & \makecell[c]{58.9} & \makecell[c]{44.2} \\
        & & \makecell[c]{Real\&Syn} & \makecell[c]{104.2 (6.6)} & \makecell[c]{58.4 (0.5)} & \makecell[c]{43.5 (0.7)} \\
    \hline
        \multirow{4}{*}{\makecell[c]{SEW-ResNet~\cite{fang2021deep}}} & \multirow{8}{*}{\makecell[c]{SNN}} & \makecell[c]{Syn} & \makecell[c]{120.3} & \makecell[c]{64.2} & \makecell[c]{49.9} \\
        & & \makecell[c]{Real} & \makecell[c]{116.8} & \makecell[c]{63.2} & \makecell[c]{49.1} \\
        & & \makecell[c]{Real\&Syn} & \makecell[c]{113.1 (3.7)} & \makecell[c]{62.5 (0.7)} & \makecell[c]{48.6 (0.5)} \\
        & & \makecell[c]{Real\&Syn + DA} & \makecell[c]{101.3 (9.9)} & \makecell[c]{56.7 (3.3)} & \makecell[c]{43.6 (1.7)} \\
    \cline{1-1}\cline{3-6}
        \multirow{4}{*}{\makecell[c]{Ours}} & & \makecell[c]{Syn} & \makecell[c]{110.7} & \makecell[c]{59.4} & \makecell[c]{45.0} \\
        & & \makecell[c]{Real} & \makecell[c]{107.2} & \makecell[c]{58.7} & \makecell[c]{44.1} \\
        & & \makecell[c]{Real\&Syn} & \makecell[c]{103.1 (4.0)} & \makecell[c]{58.4 (0.4)} & \makecell[c]{43.8 (0.3)} \\
        & & \makecell[c]{Real\&Syn + DA} & \makecell[c]{99.2 (8.0)} & \makecell[c]{55.8 (2.9)} & \makecell[c]{42.9 (1.2)} \\
    \hline \toprule 
    \end{tabular}}
    \label{tab:pose-estimation-syn}
\end{table}

\begin{figure}[t]
    \centering        
    \includegraphics[width=\columnwidth]{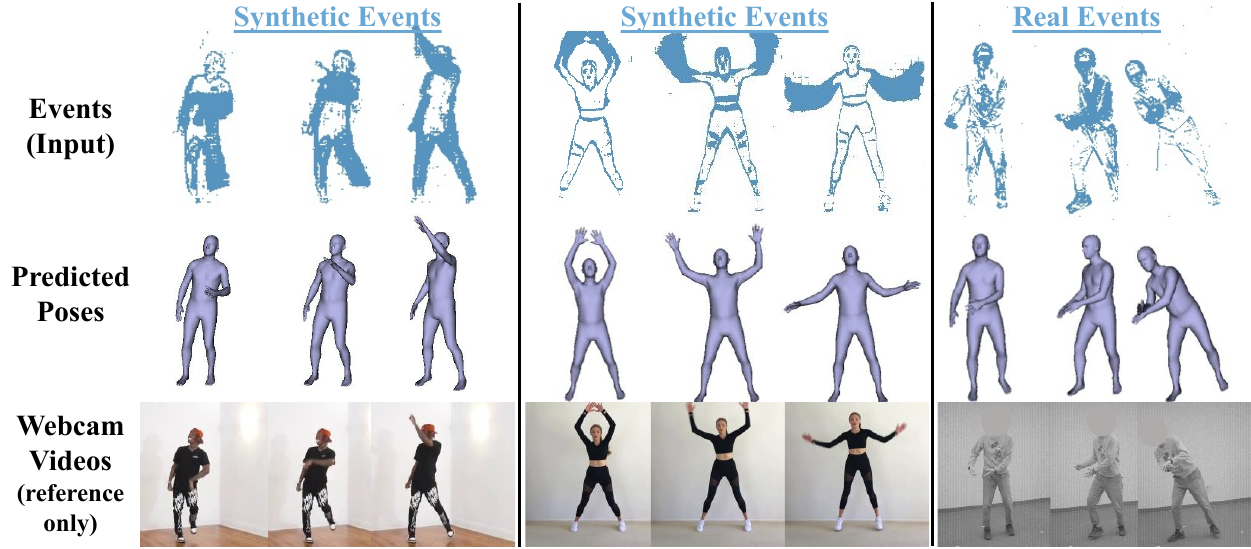}
    \caption{\textbf{Generalization ability} of our method is demonstrated by training solely on our SynEventHPD dataset and applying to unseen scenarios. The left examples utilize events synthesized from webcam videos, while the right example employs real events as input.}
    \label{fig:web_demo1}
\end{figure}

\begin{figure}[t]
    \centering
    \includegraphics[width=\columnwidth]{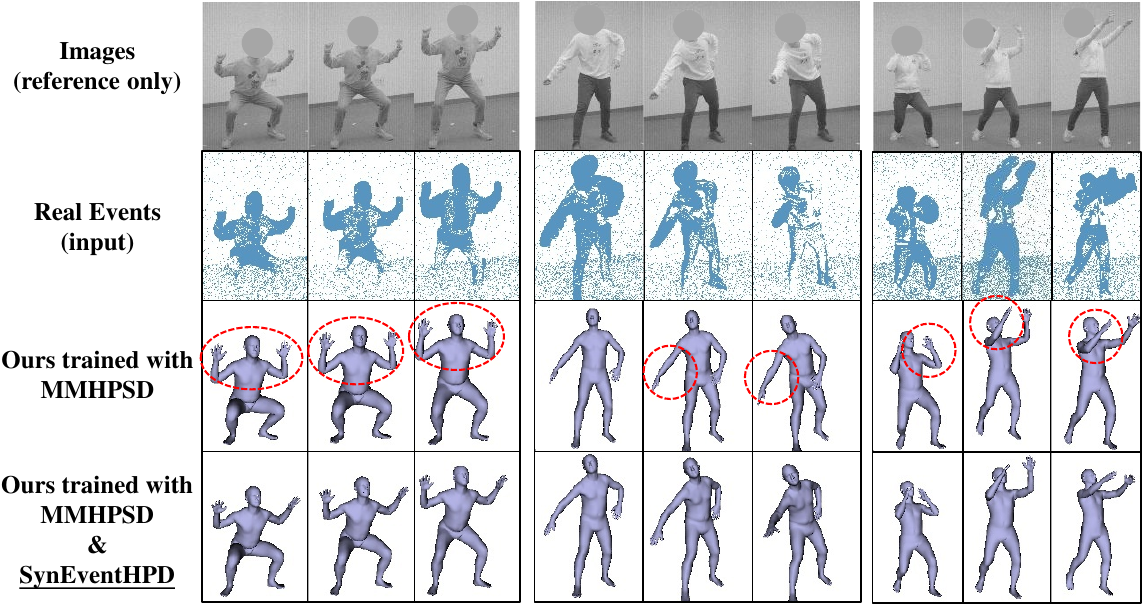}
    \caption{\textbf{Qualitative results} of models trained with and without the synthetic SynEventHPD dataset. After incorporating the SynEventHPD dataset during training, our method shows more precise pose tracking results using real events as input.}
    \label{fig:real-syn-compare}
\end{figure}

The quantitative results are displayed in Tab.~\ref{tab:pose-estimation-syn}. It is evident that, compared to models trained using the real MMHPSD train set, the pose errors are generally higher for models trained on the synthetic SynEventHPD dataset. This is largely due to the domain gap between the synthetic and real events, leading to inferior performance when training only on synthetic data and then evaluating on real data. However, after combining both real and synthetic datasets for training, all the models in Tab.~\ref{tab:pose-estimation-syn} achieve improved performance compared to training with either the real MMHPSD train set or the synthetic SynEventHPD dataset. After applying a recent domain adaptation approach~\cite{sun2023domain} during SNNs training, a noticeable improvement can be observed when compared with vanilla hybrid dataset training. This observation further demonstrates the effectiveness of our proposed SynEventHPD dataset.

The results show that models trained without the synthetic SynEventHPD dataset produce less stable and less accurate pose tracking. In contrast, incorporating the SynEventHPD dataset during training enhances generalization, leading to more precise and consistent pose estimations. It is important to note that while the additional training data consists of synthetic events, the testing data remains real event data. This improvement highlights the effectiveness of the synthetic dataset in enhancing the model’s ability to capture motion dynamics and structural cues from event signals, ultimately improving its robustness in real-world scenarios.

Our method demonstrates generalization in Fig.~\ref{fig:web_demo1} by training exclusively on the SynEventHPD dataset and applying it to unseen scenarios. The left examples showcase pose tracking using event data synthesized from webcam videos, while the right example illustrates the model's effectiveness when applied to real event inputs. The consistency in performance across both synthetic and real event data highlights the robustness of our approach, indicating its potential for real-world applications even when trained solely on synthetic data.

\subsection{Ablation Study}\label{sec:ablation}
\textbf{Spiking Attention Module Design.} Our spiking attention design differs from the Spikformer~\cite{zhou2022spikformer} in two key aspects: binary or real-valued representation of values $\mathbf{V}$ and the attention score function in spiking attention. For a fair comparison, we examine the impact of using binary or real $\mathbf{V}$ with two distinct score functions, \ie, Dot-product and our proposed normalized Hamming similarity, in both approaches. Tab.~\ref{tab:ablation-architecture} shows the PEL-MPJPE of these configurations, with energy consumption reported in the bracket. Notably, compared to designs from prior work~\cite{zhou2022spikformer}, which use binary $\mathbf{V}$ and dot-product similarity, our proposed design significantly reduces the pose tracking error (PEL-MPJPE) from 62.0 to 58.7, while increasing energy cost by only 2.5\%. This increase is still substantially lower than the energy costs associated with ANN methods. Importantly, our method allows for flexibility in spiking attention module design, depending on the desired trade-off between performance and energy consumption. A similar trend is observed for Spikformer~\cite{zhou2022spikformer} when it is replaced with our proposed spiking attention design. Additionally, with the same design, our approach consistently outperforms Spikformer, possibly due to our spatiotemporal attention over the global space-time domain while Spikformer applies over local patches, similar to ViT~\cite{dosovitskiy2020image}.

\textbf{Transformer Hidden Dimension $C_k$.} The results in Tab.~\ref{tab:ablation-architecture} indicate that setting $C_k>1024$ does not notably improve performance for a training set of 180K samples in our experiments. Conversely, it introduces a substantial increase in parameters and a heightened risk of overfitting, \eg, $C_k=4096$. The risk is mitigated when using a larger model, such as SEW-ResNet101 with 6-layer spiking attention, which achieves a PEL-MPJPE of 58.0, compared to SEW-ResNet50 with 6-layer spiking attention, which yields 58.9. This result demonstrates that a deeper backbone, combined with additional attention layers, improves spatiotemporal feature extraction, leading to more accurate pose estimation.

\textbf{Number of Attention Layers $L$.} Spikformer follows the architecture of ViT~\cite{dosovitskiy2020image} without a backbone, we configure it with $3L$ layers for a fair comparison. The addition of 1 or 2 layers of spiking attention significantly enhances performance. However, employing 6 layers leads to a reduction in performance due to the model's large parameter count of 87.4M, indicating a higher risk of overfitting.

\textbf{SEW-ResNet (SR) Backbone.} The results in Tab.~\ref{tab:ablation-architecture} show a clear trend: increasing both the backbone capacity and the number of attention layers generally improves pose estimation accuracy. Among the SEW-ResNet34 models, adding a second attention layer reduces the PEL-MPJPE from 59.7 to 59.0, indicating the benefit of deeper attention mechanisms. When upgrading to SEW-ResNet50, the 2-layer attention model achieves 58.7, while increasing to 4-layer attention slightly improves performance to 58.6. The best results are obtained with SEW-ResNet101, where 4-layer attention achieves 58.2, and 6-layer attention further reduces the error to 58.0, demonstrating that a deeper backbone combined with more attention layers enhances spatiotemporal feature extraction. However, the diminishing returns suggest a trade-off between model complexity and performance gain, as each spiking attention layer increases the parameter count by approximately 20M.

\textbf{Score functions in Spiking Spatiotemporal Transformer.} We compare the proposed normalized Hamming similarity with three commonly used score functions as detailed below:
\begin{equation}\footnotesize
    \begin{aligned}
        \text{Normalized Hamming similarity}\quad 1 - \frac{1}{C_k}\sum_{c=1}^{C_k} \mathds{1}(\mathbf{s}_{ic}^q \neq \mathbf{s}_{jc}^k), \nonumber
    \end{aligned}
\end{equation}
\begin{equation}\footnotesize
    \begin{aligned}
        \text{Scaled dot-product similarity}\quad \frac{1}{\sqrt{C_k}}\sum_{c=1}^{C_k} \mathbf{s}_{ic}^q\cdot\mathbf{s}_{jc}^k, \nonumber
    \end{aligned}
\end{equation}
\begin{equation}\footnotesize
    \begin{aligned}
        \text{Normalized Euclidean similarity}\quad 1 - \frac{1}{C_k}\sum_{c=1}^{C_k}(\mathbf{s}_{ic}^q - \mathbf{s}_{jc}^k)^2, \nonumber
    \end{aligned}
\end{equation}
\begin{equation}\footnotesize
    \begin{aligned}
        \text{Normalized Manhattan similarity}\quad 1 - \frac{1}{C_k}\sum_{c=1}^{C_k}|\mathbf{s}_{ic}^q - \mathbf{s}_{jc}^k|. \nonumber
    \end{aligned}
\end{equation}
Tab.~\ref{tab:ablation-architecture} presents the PEL-MPJPE results of four different score functions in the spiking spatiotemporal transformer. Among them, the Normalized Hamming score function achieves the lowest error of 58.7, demonstrating its superior effectiveness in capturing spatiotemporal relationships in event-based pose estimation. In comparison, the Scaled Dot-Product, Normalized Euclidean, and Normalized Manhattan score functions result in higher errors of 62.8, 61.7, and 62.3, respectively. These findings suggest that the Normalized Hamming function better preserves the sparsity and temporal dynamics of spiking neural networks, leading to more accurate pose estimation.

\begin{table}[t!]
    \centering
    \caption{\textbf{Ablation Study.} We report the PEL-MPJPE~$\downarrow$ from models ($T=8$) on the MMHPSD dataset. Results marked with \underline{\textbf{underline}} denote the outcomes of the original setting in Tab.~\ref{tab:pose-estimation}. Values in parentheses represent the energy consumption in joules.}
    \setlength{\tabcolsep}{1mm}
    \renewcommand{\arraystretch}{1.3}
    \resizebox{\columnwidth}{!}{
    \begin{tabular}{@{}|c|c|c|@{}p{2.0cm}@{}p{2.0cm}|@{}}
    \bottomrule \hline
        \multicolumn{3}{|l|}{\makecell[c]{}} & \makecell[c]{Spikformer~\cite{zhou2022spikformer}} & \makecell[c]{Ours} \\
    \hline
        \multirow{4}{*}{\makecell[c]{Spiking\\Attention\\Module\\Design}} & \multirow{2}{*}{\makecell[c]{Binary $\mathbf{V}$}} & \makecell[l]{Dot-prod.} & \makecell[c]{\underline{\textbf{62.5}} (0.0118)} & \makecell[c]{62.0 (0.00806)} \\
        & & \makecell[l]{Hamming} & \makecell[c]{59.9 (0.0121)} & \makecell[c]{59.3 (0.00820)} \\
    \cline{2-5}
        & \multirow{2}{*}{\makecell[c]{Real $\mathbf{V}$}} & \makecell[l]{Dot-prod.} & \makecell[c]{61.8 (0.0121)} & \makecell[c]{61.4 (0.00821)} \\
        & & \makecell[l]{Hamming} & \makecell[c]{59.5 (0.0124)} & \makecell[c]{\underline{\textbf{58.7}} (0.00826)} \\
    \hline
        \multirow{5}{*}{\makecell[c]{Transformer\\Hidden\\Dimension\\$C_k$}} & \multicolumn{2}{|c|}{\makecell[l]{\textbf{256} (30.1M params)}}  & \makecell[c]{70.3} & \makecell[c]{66.2} \\
        & \multicolumn{2}{|c|}{\makecell[l]{\textbf{512} (36.4M params)}}  & \makecell[c]{65.8} & \makecell[c]{62.1} \\
        & \multicolumn{2}{|c|}{\makecell[l]{\textbf{1024} (47.7M params)}}  & \makecell[c]{\underline{\textbf{62.5}}} & \makecell[c]{\underline{\textbf{58.7}}} \\
        & \multicolumn{2}{|c|}{\makecell[l]{\textbf{2048} (74.2M params)}}  & \makecell[c]{62.4} & \makecell[c]{58.5} \\
        & \multicolumn{2}{|c|}{\makecell[l]{\textbf{4096} (124.5M params)}}  & \makecell[c]{-} & \makecell[c]{59.0} \\
    \hline
        \multirow{5}{*}{\makecell[c]{\# of\\Attention\\Layers\\$L$}} & \multicolumn{2}{|c|}{\makecell[l]{\textbf{0} (25.8M params)}}  & \makecell[c]{-} & \makecell[c]{63.2} \\
        & \multicolumn{2}{|c|}{\makecell[l]{\textbf{1} (36.4M params)}}  & \makecell[c]{67.9} & \makecell[c]{60.5} \\
        & \multicolumn{2}{|c|}{\makecell[l]{\textbf{2} (47.7M params)}}  & \makecell[c]{\underline{\textbf{62.5}}} & \makecell[c]{\underline{\textbf{58.7}}} \\
        & \multicolumn{2}{|c|}{\makecell[l]{\textbf{4} (67.6M params)}}  & \makecell[c]{60.0} & \makecell[c]{58.6} \\
        & \multicolumn{2}{|c|}{\makecell[l]{\textbf{6} (87.4M params)}}  & \makecell[c]{60.2} & \makecell[c]{58.9} \\
    \hline
        \multirow{4}{*}{\makecell[c]{Score\\Function}} & \multicolumn{2}{|c|}{\makecell[l]{Normalized Hamming}}  & \makecell[c]{\underline{\textbf{62.5}}} & \makecell[c]{\underline{\textbf{58.7}}} \\
        & \multicolumn{2}{|c|}{\makecell[l]{Scaled Dot-Product}}  & \makecell[c]{64.7} & \makecell[c]{62.8} \\
        & \multicolumn{2}{|c|}{\makecell[l]{Normalized Euclidean}}  & \makecell[c]{63.5} & \makecell[c]{61.7} \\
        & \multicolumn{2}{|c|}{\makecell[l]{Normalized Manhattan}}  & \makecell[c]{63.9} & \makecell[c]{62.3} \\
    \hline
        \multirow{6}{*}{\makecell[c]{SEW-ResNet\\(SR)\\Backbone}} & \multicolumn{2}{|c|}{\makecell[l]{SR34 + 1-layer Att.}}  & \makecell[c]{-} & \makecell[c]{59.7} \\
        & \multicolumn{2}{|c|}{\makecell[l]{SR34 + 2-layer Att.}}  & \makecell[c]{-} & \makecell[c]{59.0} \\
        & \multicolumn{2}{|c|}{\makecell[l]{SR50 + 2-layer Att.}}  & \makecell[c]{-} & \makecell[c]{\underline{\textbf{58.7}}} \\
        & \multicolumn{2}{|c|}{\makecell[l]{SR50 + 4-layer Att.}}  & \makecell[c]{-} & \makecell[c]{58.6} \\
        & \multicolumn{2}{|c|}{\makecell[l]{SR101 + 4-layer Att.}}  & \makecell[c]{-} & \makecell[c]{58.2} \\
        & \multicolumn{2}{|c|}{\makecell[l]{SR101 + 6-layer Att.}}  & \makecell[c]{-} & \makecell[c]{58.0} \\
    \hline \toprule 
    \end{tabular}}
    \label{tab:ablation-architecture}
\end{table}

\subsection{Discussion}
\label{sec:limitation}

Our approach tracks 3d human poses from events, while event cameras can only generate events for moving objects. A limitation arises from the minimal movement of body parts, leading to insufficient events for those static regions. Our proposed spiking spatiotemporal attention can alleviate this limitation to some extent by integrating bi-directional contextual information for pose tracking over time. Another possible solution involves accumulating events for a longer time, which increases the probability to capture the contextual events information of full-body poses. These accumulated events can subsequently be processed at different temporal scales of granularity when feeding into our model for pose tracking.

\begin{figure}[t]
    \centering
    \includegraphics[width=\columnwidth]{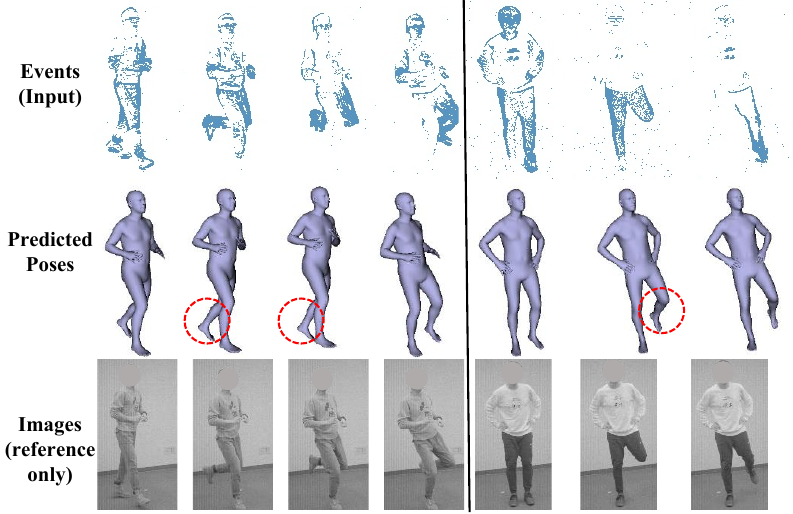}
    \caption{\textbf{Failure cases.} Owing to the presence of body part occlusion and also the absence of temporal context, our method struggles to accurately estimate poses, as indicated by the red circles.}
    \label{fig:failure-case}
\end{figure}

\begin{figure}[t]
    \centering
    \includegraphics[width=\columnwidth]{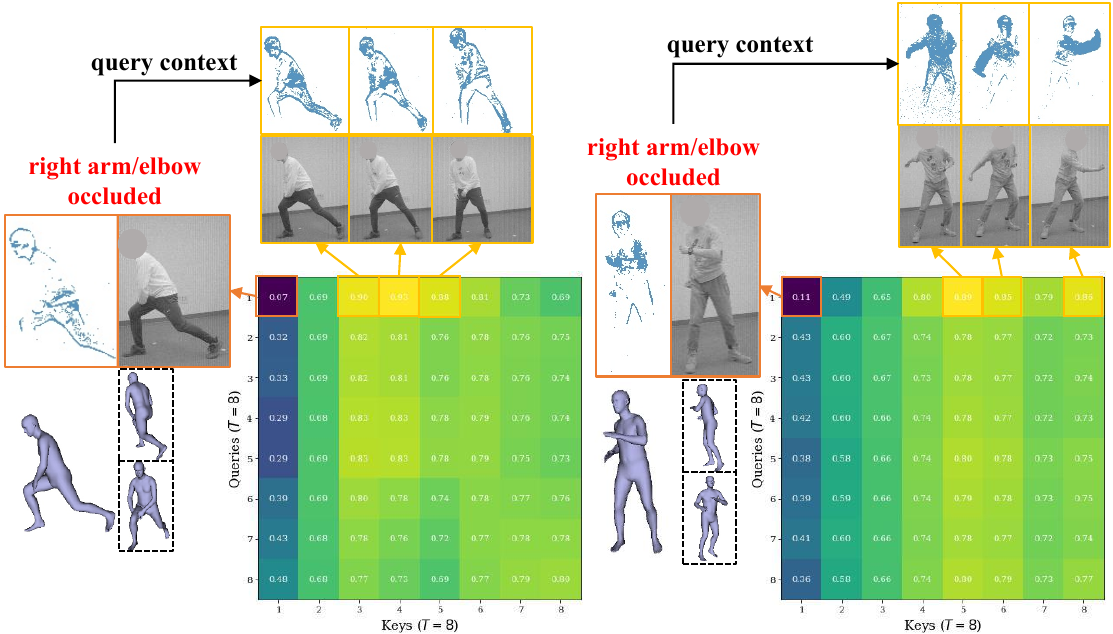}
    \caption{\textbf{Visualization of attention score maps.} Our approach determines accurate poses for obscured body parts at an early stage using spiking spatiotemporal attention, in which the query at $t=1$ places significantly greater emphasis on features from later time steps.}
    \label{fig:att-map}
\end{figure}

Events fundamentally represent 2D signals that lack robust 3D information, which constrains pose estimation for occluded body parts. Fig.~\ref{fig:failure-case} illustrates failure cases where poses are inaccurately estimated from the events. These inaccuracies are primarily attributed to body part occlusion. Additionally, the lack of temporal context exacerbates this issue, preventing the model from effectively leveraging information from previous or subsequent frames to compensate for missing or obscured data. However, when temporal context provides relevant clues for occluded body parts, our method assigns higher attention scores to those specific time steps, effectively capturing the missing information. To illustrate this, we provide visualizations of attention score maps in Fig.~\ref{fig:att-map}. Our approach enhances pose estimation for obscured body parts at an early stage by leveraging spiking spatiotemporal attention, where the query at t=1 prioritizes features from later time steps to infer occluded body parts more accurately. In this context, introducing multi-view events for efficient pose tracking may offer a robust solution by providing complementary perspectives that mitigate the impact of occlusions. By integrating event streams from multiple angles, the system can enhance depth perception and improve spatial consistency, thereby reducing ambiguities in pose estimation. Furthermore, leveraging spiking spatiotemporal attention across multi-view inputs can enable more efficient feature alignment and information fusion, ensuring a more stable and accurate tracking performance even in challenging scenarios with severe occlusions.

Future work could focus on developing an end-to-end SNN-based pose estimation framework that directly outputs 3D joint positions instead of relying on the parametric SMPL pose and shape model. While SMPL provides a more comprehensive representation of the human body, its axis-angle parameters introduce challenges in model training and generalization, making direct 3D pose estimation a more efficient alternative. Moving beyond feature encoding, an end-to-end SNN framework could jointly learn spatial-temporal patterns, enabling more robust and adaptive pose estimation. By leveraging the event-driven nature of SNNs, the framework could efficiently process sparse, asynchronous data, reducing computational costs while improving real-time performance. This would be particularly beneficial for downstream tasks such as real-time object tracking, behavior analysis, and action recognition, where efficiency and responsiveness are critical. Furthermore, leveraging 3D pose representation would facilitate the development of specialized, high-efficiency SNN approaches for integrating multi-view event streams or incorporating traditional full-frame images, ultimately enhancing human pose tracking in multi-view scenarios. Exploring hardware acceleration strategies, such as deploying the SNN framework on neuromorphic chips, could further optimize power efficiency and inference speed, making it a viable solution for real-time applications in robotics, augmented reality, and autonomous systems.

\section{Conclusion}
In conclusion, this work introduces the first SNN-based approach for event-driven 3D human pose tracking, leveraging event sparsity for exceptional computational and energy efficiency. The proposed Spiking Spatiotemporal Transformer overcomes the temporal dependency limitations in SNNs and ensures robust similarity measurement in spiking attention. To support research in this domain, we provide SynEventHPD, a large-scale dataset with 45.72 hours of event streams, vastly surpassing existing datasets. Extensive experiments demonstrate that our method outperforms SOTA ANN-based approaches with significantly lower computational cost while achieving superior accuracy compared to existing SNN models, underscoring the effectiveness of our framework.

\section*{Acknowledgment}
This work was partially supported by a grant from the  grants from National Natural Science Foundation of China (62372441, U22A2034), in part by Guangdong Basic and
Applied Basic Research Foundation (2023A1515030268),  and in part by Shenzhen Science and Technology Program (Grant No. RCYX20231211090127030, JSGG20220831105002004). 



\bibliographystyle{IEEEtran}
\bibliography{IEEEabrv}

\begin{thebibliography}{10}
\providecommand{\url}[1]{#1}
\csname url@samestyle\endcsname
\providecommand{\newblock}{\relax}
\providecommand{\bibinfo}[2]{#2}
\providecommand{\BIBentrySTDinterwordspacing}{\spaceskip=0pt\relax}
\providecommand{\BIBentryALTinterwordstretchfactor}{4}
\providecommand{\BIBentryALTinterwordspacing}{\spaceskip=\fontdimen2\font plus
\BIBentryALTinterwordstretchfactor\fontdimen3\font minus \fontdimen4\font\relax}
\providecommand{\BIBforeignlanguage}[2]{{%
\expandafter\ifx\csname l@#1\endcsname\relax
\typeout{** WARNING: IEEEtran.bst: No hyphenation pattern has been}%
\typeout{** loaded for the language `#1'. Using the pattern for}%
\typeout{** the default language instead.}%
\else
\language=\csname l@#1\endcsname
\fi
#2}}
\providecommand{\BIBdecl}{\relax}
\BIBdecl

\bibitem{gai2023spatiotemporal}
D.~Gai, R.~Feng, W.~Min, X.~Yang, P.~Su, Q.~Wang, and Q.~Han, ``Spatiotemporal learning transformer for video-based human pose estimation,'' \emph{IEEE TCSVT}, 2023.

\bibitem{zhou2024dual}
L.~Zhou, Y.~Chen, and J.~Wang, ``Dual-path transformer for 3d human pose estimation,'' \emph{IEEE TCSVT}, 2024.

\bibitem{tang2024ftcm}
Z.~Tang, Y.~Hao, J.~Li, and R.~Hong, ``Ftcm: Frequency-temporal collaborative module for efficient 3d human pose estimation in video,'' \emph{IEEE TCSVT}, 2024.

\bibitem{shi2024identify}
C.~Shi, Y.~Li, N.~Song, B.~Wei, Y.~Zhang, W.~Li, and J.~Jin, ``Identifying light interference in event-based vision,'' \emph{IEEE TCSVT}, 2024.

\bibitem{liu2024event}
X.~Liu, J.~Li, J.~Shi, X.~Fan, Y.~Tian, and D.~Zhao, ``Event-based monocular depth estimation with recurrent transformers,'' \emph{IEEE TCSVT}, 2024.

\bibitem{humanMotionKanazawa19}
A.~Kanazawa, J.~Y. Zhang, P.~Felsen, and J.~Malik, ``Learning 3d human dynamics from video,'' in \emph{CVPR}, 2019.

\bibitem{kocabas2019vibe}
M.~Kocabas, N.~Athanasiou, and M.~J. Black, ``Vibe: Video inference for human body pose and shape estimation,'' in \emph{CVPR}, 2020.

\bibitem{gallego2019event}
G.~Gallego, T.~Delbruck, G.~Orchard, C.~Bartolozzi, B.~Taba, A.~Censi, S.~Leutenegger, A.~Davison, J.~Conradt, K.~Daniilidis, and D.~Scaramuzza, ``Event-based vision: A survey,'' \emph{IEEE TPAMI}, 2022.

\bibitem{zhou2023spikingformer}
C.~Zhou, L.~Yu, Z.~Zhou, H.~Zhang, Z.~Ma, H.~Zhou, and Y.~Tian, ``Spikingformer: Spike-driven residual learning for transformer-based spiking neural network,'' \emph{arXiv preprint arXiv:2304.11954}, 2023.

\bibitem{zhang2021object}
J.~Zhang, X.~Yang, Y.~Fu, X.~Wei, B.~Yin, and B.~Dong, ``Object tracking by jointly exploiting frame and event domain,'' in \emph{ICCV}, 2021.

\bibitem{zhang2022spiking}
J.~Zhang, B.~Dong, H.~Zhang, J.~Ding, F.~Heide, B.~Yin, and X.~Yang, ``Spiking transformers for event-based single object tracking,'' in \emph{CVPR}, 2022.

\bibitem{kim2022ev}
J.~Kim, I.~Hwang, and Y.~M. Kim, ``Ev-tta: Test-time adaptation for event-based object recognition,'' in \emph{CVPR}, 2022.

\bibitem{fang2021deep}
W.~Fang, Z.~Yu, Y.~Chen, T.~Huang, T.~Masquelier, and Y.~Tian, ``Deep residual learning in spiking neural networks,'' \emph{NeurIPS}, 2021.

\bibitem{rebecq2018emvs}
H.~Rebecq, G.~Gallego, E.~Mueggler, and D.~Scaramuzza, ``Emvs: Event-based multi-view stereo—3d reconstruction with an event camera in real-time,'' \emph{IJCV}, 2018.

\bibitem{zhang2022discrete}
K.~Zhang, K.~Che, J.~Zhang, J.~Cheng, Z.~Zhang, Q.~Guo, and L.~Leng, ``Discrete time convolution for fast event-based stereo,'' in \emph{CVPR}, 2022.

\bibitem{calabrese2019dhp19}
E.~Calabrese, G.~Taverni, C.~Awai~Easthope, S.~Skriabine, F.~Corradi, L.~Longinotti, K.~Eng, and T.~Delbruck, ``Dhp19: Dynamic vision sensor 3d human pose dataset,'' in \emph{CVPR Workshop}, 2019.

\bibitem{xu2020eventcap}
L.~Xu, W.~Xu, V.~Golyanik, M.~Habermann, L.~Fang, and C.~Theobalt, ``Eventcap: Monocular 3d capture of high-speed human motions using an event camera,'' in \emph{CVPR}, 2020.

\bibitem{zou2021eventhpe}
S.~Zou, C.~Guo, X.~Zuo, S.~Wang, P.~Wang, X.~Hu, S.~Chen, M.~Gong, and L.~Cheng, ``Eventhpe: Event-based 3d human pose and shape estimation,'' in \emph{ICCV}, 2021.

\bibitem{rudnev2021eventhands}
V.~Rudnev, V.~Golyanik, J.~Wang, H.-P. Seidel, F.~Mueller, M.~Elgharib, and C.~Theobalt, ``Eventhands: real-time neural 3d hand pose estimation from an event stream,'' in \emph{ICCV}, 2021.

\bibitem{scarpellini2021lifting}
G.~Scarpellini, P.~Morerio, and A.~Del~Bue, ``Lifting monocular events to 3d human poses,'' in \emph{CVPR}, 2021.

\bibitem{he2016deep}
K.~He, X.~Zhang, S.~Ren, and J.~Sun, ``Deep residual learning for image recognition,'' in \emph{CVPR}, 2016.

\bibitem{vaswani2017attention}
A.~Vaswani, N.~Shazeer, N.~Parmar, J.~Uszkoreit, L.~Jones, A.~N. Gomez, {\L}.~Kaiser, and I.~Polosukhin, ``Attention is all you need,'' \emph{NeurIPS}, 2017.

\bibitem{wang2020eventsr}
L.~Wang, T.-K. Kim, and K.-J. Yoon, ``Eventsr: From asynchronous events to image reconstruction, restoration, and super-resolution via end-to-end adversarial learning,'' in \emph{CVPR}, 2020.

\bibitem{fang2021incorporating}
W.~Fang, Z.~Yu, Y.~Chen, T.~Masquelier, T.~Huang, and Y.~Tian, ``Incorporating learnable membrane time constant to enhance learning of spiking neural networks,'' in \emph{ICCV}, 2021.

\bibitem{sun2022ess}
Z.~Sun, N.~Messikommer, D.~Gehrig, and D.~Scaramuzza, ``Ess: Learning event-based semantic segmentation from still images,'' in \emph{ECCV}, 2022.

\bibitem{gehrig2019end}
D.~Gehrig, A.~Loquercio, K.~G. Derpanis, and D.~Scaramuzza, ``End-to-end learning of representations for asynchronous event-based data,'' in \emph{ICCV}, 2019.

\bibitem{gehrig2020video}
D.~Gehrig, M.~Gehrig, J.~Hidalgo-Carri{\'o}, and D.~Scaramuzza, ``Video to events: Recycling video datasets for event cameras,'' in \emph{CVPR}, 2020.

\bibitem{rueckauer2017conversion}
B.~Rueckauer, I.-A. Lungu, Y.~Hu, M.~Pfeiffer, and S.-C. Liu, ``Conversion of continuous-valued deep networks to efficient event-driven networks for image classification,'' \emph{Frontiers in neuroscience}, 2017.

\bibitem{deng2021optimal}
S.~Deng and S.~Gu, ``Optimal conversion of conventional artificial neural networks to spiking neural networks,'' in \emph{ICLR}, 2021.

\bibitem{li2021differentiable}
Y.~Li, Y.~Guo, S.~Zhang, S.~Deng, Y.~Hai, and S.~Gu, ``Differentiable spike: Rethinking gradient-descent for training spiking neural networks,'' \emph{NeurIPS}, 2021.

\bibitem{yao2022glif}
X.~Yao, F.~Li, Z.~Mo, and J.~Cheng, ``Glif: A unified gated leaky integrate-and-fire neuron for spiking neural networks,'' in \emph{NeurIPS}, 2022.

\bibitem{yao2023attention}
M.~Yao, G.~Zhao, H.~Zhang, Y.~Hu, L.~Deng, Y.~Tian, B.~Xu, and G.~Li, ``Attention spiking neural networks,'' \emph{IEEE TPAMI}, 2023.

\bibitem{zhou2022spikformer}
Z.~Zhou, Y.~Zhu, C.~He, Y.~Wang, S.~Yan, Y.~Tian, and L.~Yuan, ``Spikformer: When spiking neural network meets transformer,'' in \emph{ICLR}, 2022.

\bibitem{zhang2022spikedepth}
J.~Zhang, L.~Tang, Z.~Yu, J.~Lu, and T.~Huang, ``Spike transformer: Monocular depth estimation for spiking camera,'' in \emph{ECCV}, 2022.

\bibitem{carion2020end}
N.~Carion, F.~Massa, G.~Synnaeve, N.~Usunier, A.~Kirillov, and S.~Zagoruyko, ``End-to-end object detection with transformers,'' in \emph{ECCV}, 2020.

\bibitem{zhou2016sparseness}
X.~Zhou, M.~Zhu, S.~Leonardos, K.~G. Derpanis, and K.~Daniilidis, ``Sparseness meets deepness: 3d human pose estimation from monocular video,'' in \emph{CVPR}, 2016.

\bibitem{wang20193d}
K.~Wang, L.~Lin, C.~Jiang, C.~Qian, and P.~Wei, ``3d human pose machines with self-supervised learning,'' \emph{IEEE TPAMI}, 2020.

\bibitem{loper2015smpl}
M.~Loper, N.~Mahmood, J.~Romero, G.~Pons-Moll, and M.~J. Black, ``Smpl: A skinned multi-person linear model,'' \emph{ACM TOG}, 2015.

\bibitem{kanazawa2018end}
A.~Kanazawa, M.~J. Black, D.~W. Jacobs, and J.~Malik, ``End-to-end recovery of human shape and pose,'' in \emph{CVPR}, 2018.

\bibitem{gallego2017event}
G.~Gallego, J.~E. Lund, E.~Mueggler, H.~Rebecq, T.~Delbruck, and D.~Scaramuzza, ``Event-based, 6-dof camera tracking from photometric depth maps,'' \emph{IEEE TPAMI}, 2017.

\bibitem{gehrig2018asynchronous}
D.~Gehrig, H.~Rebecq, G.~Gallego, and D.~Scaramuzza, ``Asynchronous, photometric feature tracking using events and frames,'' in \emph{ECCV}, 2018.

\bibitem{hagenaars2021self}
J.~Hagenaars, F.~Paredes-Vall{\'e}s, and G.~De~Croon, ``Self-supervised learning of event-based optical flow with spiking neural networks,'' \emph{NeurIPS}, 2021.

\bibitem{sun2022event}
L.~Sun, C.~Sakaridis, J.~Liang, Q.~Jiang, K.~Yang, P.~Sun, Y.~Ye, K.~Wang, and L.~V. Gool, ``Event-based fusion for motion deblurring with cross-modal attention,'' in \emph{ECCV}, 2022.

\bibitem{li2017cifar10}
H.~Li, H.~Liu, X.~Ji, G.~Li, and L.~Shi, ``Cifar10-dvs: an event-stream dataset for object classification,'' \emph{Frontiers in neuroscience}, 2017.

\bibitem{lin2021imagenet}
Y.~Lin, W.~Ding, S.~Qiang, L.~Deng, and G.~Li, ``Es-imagenet: A million event-stream classification dataset for spiking neural networks,'' \emph{Frontiers in neuroscience}, 2021.

\bibitem{yao2021temporal}
M.~Yao, H.~Gao, G.~Zhao, D.~Wang, Y.~Lin, Z.~Yang, and G.~Li, ``Temporal-wise attention spiking neural networks for event streams classification,'' in \emph{ICCV}, 2021.

\bibitem{zhang2018adversarial}
K.~Zhang, W.~Luo, Y.~Zhong, L.~Ma, W.~Liu, and H.~Li, ``Adversarial spatio-temporal learning for video deblurring,'' \emph{IEEE TIP}, 2018.

\bibitem{zhang2022enhanced}
K.~Zhang, D.~Li, W.~Luo, W.~Ren, and W.~Liu, ``Enhanced spatio-temporal interaction learning for video deraining: faster and better,'' \emph{IEEE TPAMI}, 2022.

\bibitem{liang2022self}
H.~Liang, N.~Quader, Z.~Chi, L.~Chen, P.~Dai, J.~Lu, and Y.~Wang, ``Self-supervised spatiotemporal representation learning by exploiting video continuity,'' in \emph{AAAI}, 2022.

\bibitem{gu2024context}
X.~Gu, H.~Fan, Y.~Huang, T.~Luo, and L.~Zhang, ``Context-guided spatio-temporal video grounding,'' in \emph{CVPR}, 2024.

\bibitem{horowitz20141}
M.~Horowitz, ``1.1 computing's energy problem (and what we can do about it),'' in \emph{ISSCC}, 2014.

\bibitem{jacques2013robust}
L.~Jacques, J.~N. Laska, P.~T. Boufounos, and R.~G. Baraniuk, ``Robust 1-bit compressive sensing via binary stable embeddings of sparse vectors,'' \emph{IEEE Transactions on Information Theory}, 2013.

\bibitem{yi2015binary}
X.~Yi, C.~Caramanis, and E.~Price, ``Binary embedding: Fundamental limits and fast algorithm,'' in \emph{ICML}, 2015.

\bibitem{zhou2019continuity}
Y.~Zhou, C.~Barnes, J.~Lu, J.~Yang, and H.~Li, ``On the continuity of rotation representations in neural networks,'' in \emph{CVPR}, 2019.

\bibitem{ionescu2013human3}
C.~Ionescu, D.~Papava, V.~Olaru, and C.~Sminchisescu, ``Human3. 6m: Large scale datasets and predictive methods for 3d human sensing in natural environments,'' \emph{IEEE TPAMI}, 2013.

\bibitem{mahmood2019amass}
N.~Mahmood, N.~Ghorbani, N.~F. Troje, G.~Pons-Moll, and M.~J. Black, ``Amass: Archive of motion capture as surface shapes,'' in \emph{ICCV}, 2019.

\bibitem{zou2022human}
S.~Zou, X.~Zuo, S.~Wang, Y.~Qian, C.~Guo, and L.~Cheng, ``Human pose and shape estimation from single polarization images,'' \emph{IEEE TMM}, 2022.

\bibitem{SpikingJelly}
W.~Fang, Y.~Chen, J.~Ding, D.~Chen, Z.~Yu, H.~Zhou, Y.~Tian, and other contributors, ``Spikingjelly,'' \url{https://github.com/fangwei123456/spikingjelly}, 2020, accessed: 2025-01-16.

\bibitem{wei2022capturing}
W.-L. Wei, J.-C. Lin, T.-L. Liu, and H.-Y.~M. Liao, ``Capturing humans in motion: temporal-attentive 3d human pose and shape estimation from monocular video,'' in \emph{CVPR}, 2022.

\bibitem{liu2021swin}
Z.~Liu, Y.~Lin, Y.~Cao, H.~Hu, Y.~Wei, Z.~Zhang, S.~Lin, and B.~Guo, ``Swin transformer: Hierarchical vision transformer using shifted windows,'' in \emph{ICCV}, 2021.

\bibitem{shen2023global}
X.~Shen, Z.~Yang, X.~Wang, J.~Ma, C.~Zhou, and Y.~Yang, ``Global-to-local modeling for video-based 3d human pose and shape estimation,'' in \emph{CVPR}, 2023.

\bibitem{pei2019towards}
J.~Pei, L.~Deng, S.~Song, M.~Zhao, Y.~Zhang, S.~Wu, G.~Wang, Z.~Zou, Z.~Wu, W.~He \emph{et~al.}, ``Towards artificial general intelligence with hybrid tianjic chip architecture,'' \emph{Nature}, 2019.

\bibitem{dosovitskiy2020image}
A.~Dosovitskiy, L.~Beyer, A.~Kolesnikov, D.~Weissenborn, X.~Zhai, T.~Unterthiner, M.~Dehghani, M.~Minderer, G.~Heigold, S.~Gelly \emph{et~al.}, ``An image is worth 16x16 words: Transformers for image recognition at scale,'' in \emph{ICLR}, 2020.

\bibitem{sun2023domain}
T.~Sun, C.~Lu, and H.~Ling, ``Domain adaptation with adversarial training on penultimate activations,'' in \emph{AAAI}, 2023.

\end{thebibliography}
\end{document}